%% file: main.tex
\documentclass{article}

\usepackage{microtype}
\usepackage{graphicx}
\usepackage{subfigure}
\usepackage{booktabs} 

\usepackage{hyperref}

\usepackage[accepted]{icml2024}

\usepackage{amsmath}
\usepackage{amssymb}
\usepackage{mathtools}
\usepackage{amsthm}

\usepackage[capitalize,noabbrev]{cleveref}

\theoremstyle{plain}

\theoremstyle{definition}

\theoremstyle{remark}

\usepackage[textsize=tiny]{todonotes}
\usepackage{enumitem}

\usepackage{xspace}
\newcommand{\eg}{{\em e.g.}\xspace}
\newcommand{\ie}{{\em i.e.}\xspace}

\usepackage[normalem]{ulem}

\icmltitlerunning{Quality Diversity through Human Feedback: 
Towards Open-Ended Diversity-Driven Optimization}

\begin{document}

\twocolumn[
  \icmltitle{Quality Diversity through Human Feedback: \\
    Towards Open-Ended Diversity-Driven Optimization}



  \icmlsetsymbol{stability}{*}

  \begin{icmlauthorlist}
    \icmlauthor{Li Ding}{umass}
    \icmlauthor{Jenny Zhang}{ubc,vector}
    \icmlauthor{Jeff Clune}{ubc,vector,cifar}
    \icmlauthor{Lee Spector}{amherst,umass}
    \icmlauthor{Joel Lehman}{stochastic,stability}
  \end{icmlauthorlist}

  \icmlaffiliation{umass}{Manning College of Information \& Computer Sciences, University of Massachusetts Amherst}
  \icmlaffiliation{ubc}{Department of Computer Science, University of British Columbia}
  \icmlaffiliation{vector}{Vector Institute}
  \icmlaffiliation{cifar}{Canada CIFAR AI Chair}
  \icmlaffiliation{amherst}{Department of Computer Science, Amherst College}
  \icmlaffiliation{stochastic}{Stochastic Labs *Part of the work was done while the author was affiliated with Stability AI}

  \icmlcorrespondingauthor{Li Ding}{liding@umass.edu}

  \icmlkeywords{Machine Learning, ICML}

  \vskip 0.3in
]



\printAffiliationsAndNotice{}  

\begin{abstract}
  Reinforcement Learning from Human Feedback (RLHF) has shown potential in qualitative tasks where easily defined performance measures are lacking. However, there are drawbacks when RLHF is commonly used to optimize for average human preferences, especially in generative tasks that demand diverse model responses. Meanwhile, Quality Diversity (QD) algorithms excel at identifying diverse and high-quality solutions but often rely on manually crafted diversity metrics.
  This paper introduces Quality Diversity through Human Feedback (QDHF), a novel approach that progressively infers diversity metrics from human judgments of similarity among solutions, thereby enhancing the applicability and effectiveness of QD algorithms in complex and open-ended domains. Empirical studies show that QDHF significantly outperforms state-of-the-art methods in automatic diversity discovery and matches the efficacy of QD with manually crafted diversity metrics on standard benchmarks in robotics and reinforcement learning. Notably, in open-ended generative tasks, QDHF substantially enhances the diversity of text-to-image generation from a diffusion model and is more favorably received in user studies.
  We conclude by analyzing QDHF's scalability, robustness, and quality of derived diversity metrics, emphasizing its strength in open-ended optimization tasks. Code and tutorials are available at \url{https://liding.info/qdhf}.
\end{abstract}

\section{Introduction}
Foundation models such as large language models (LLMs) and text-to-image generation models in effect compress vast archives of human knowledge into powerful and flexible tools, serving as a foundation for down-stream applications~\citep{brown2020language,bommasani2021opportunities}.
Their promise includes helping individuals better meet their varied goals, such as exploring their creativity in different modalities and coming up with novel solutions.
One mechanism to build upon such foundational knowledge is reinforcement learning from human feedback (RLHF)~\citep{christiano2017deep}, which can make models both easier to use (by aligning them to human instructions), and more competent (by improving their capabilities based on human preferences).

RLHF is a relatively new paradigm, and its deployments often follow the relatively narrow recipe of maximizing a learned reward model of averaged human preferences over model responses.
This work aims to broaden that recipe to include optimizing for interesting diversity among responses, which is of practical importance for many creative applications such as text-to-image generation~\citep{rombach2022high}.
Such increased diversity can also improve optimization for complex and open-ended tasks (through improved exploration), personalization (serving individual rather than average human preference), and fairness (to offset algorithmic biases in gender, ethnicity, and more).

Diversity encourages exploration, which is essential for finding novel and effective solutions to many complex problems. Without diversity, optimization algorithms can converge prematurely, resulting in getting stuck in local optima or producing only a limited set of responses (\ie, mode collapse). Diversity-seeking is a core aspect of Quality Diversity (QD) algorithms~\citep{pugh2016quality,cully2015robots,lehman2011evolving,mouret2015illuminating}, where diversity metrics are explicitly defined and utilized to encourage the variation of solutions during optimization.

Our main idea is to derive distinct representations of what humans find interestingly different and use such diversity representations to support optimization. We introduce a novel method, Quality Diversity through Human Feedback (QDHF), which empowers QD algorithms with diversity metrics actively learned from human feedback during optimization. QDHF serves as an online, diversity-driven optimizer for open-ended tasks. It navigates the search space through the application of progressively learned diversity, which is achieved by concurrently fine-tuning inferred diversity by aligning it to human feedback in response to the discovery of novel solutions.

To illustrate its capability, we propose a generic implementation of QDHF, which is capable of formulating arbitrary diversity metrics using latent space projection, and aligning them to human feedback through contrastive learning~\citep{hadsell2006dimensionality,dosovitskiy2014discriminative}.
Experiments are conducted on benchmarks across three domains: robotics, reinforcement learning (RL), and generative modeling, demonstrating QDHF's superior performance in producing diverse, high-quality responses.

This work is inspired by the considerable benefits that learned reward models have unlocked for RLHF. Analogous to reward functions, diversity metrics are often qualitative, complex, and hard to specify.
Existing QD algorithms demonstrate proficiency in addressing complex search tasks, but their reliance on manually crafted diversity metrics restricts their applicability in real-world open-ended tasks.
QDHF aims to bridge this gap, allowing QD algorithms to easily adapt to more challenging tasks by learning diversity metrics through iterative exploration and alignment.
In summary, the main contributions of this work are:
\begin{enumerate}[topsep=0pt,parsep=0pt]
  \item Introducing QDHF and its implementation leveraging latent projection and contrastive learning.
  \item Demonstrating that QDHF mirrors the search capabilities of QD algorithms with manually crafted diversity metrics in benchmark robotics and RL tasks.
  \item Implementing QDHF in the latent space illumination (LSI) task for open-ended text-to-image generation, showing its capability in improving diffusion model to generate more diverse and preferable responses.
  \item Ablation studies on QDHF's sample efficiency, robustness, and the quality of its learned diversity metrics.
\end{enumerate}

\section{Preliminaries}
\label{sec:concepts}

This section covers the basic and most relevant aspects of QD algorithms. More detailed descriptions are included in Appendix~\ref{sec:apqd}.

\subsection{Quality Diversity}

QD algorithms effectively explore the search space by maintaining diverse high-quality solutions and using them to drive optimization.
Given a solution space $\mathcal{X}$, QD considers an objective function $J: \mathcal{X} \rightarrow \mathbb{R}$ and $k$ diversity metrics $M_i: \mathcal{X} \rightarrow \mathbb{R}$, $i=1,2,\cdots,k$.
The diversity metrics jointly form a measurement space $M(\mathcal{X})$, which quantifies the diversity of samples.
For each unique measurement in $M(\mathcal{X})$, the global objective $J^*$ of QD is to find a solution $x \in \mathcal{X}$ that has a maximum $J(x)$ among all solutions with the same diversity measurement.

Considering that the measurement space $M(\mathcal{X})$ is usually continuous, a QD algorithm will ultimately need to store an infinite number of solutions corresponding to each solution in the measurement space. One common way to mitigate this is to discretize $M(\mathcal{X})$ into an archive of $s$ cells $\{C_1, C_2, \cdots, C_s\}$, which was introduced in MAP-Elites~\citep{mouret2015illuminating,cully2015robots} and has been widely adopted. The QD objective is thus approximated by a relaxation to finding a set of solutions $\{x_i\}, i \in \{1, . . . , s\}$, and each $x_i$ is the best solution for one unique cell $C_i$. This (approximated) $J^*$ is:
\begin{equation}\label{eq:qd}
  J^* = \sum_{i=1}^{s} \max_{x\in\mathcal{X}, M(x)\in C_i} J(x).
\end{equation}

\subsection{Quality Diversity w/o Predefined Diversity Metrics}
\label{sec:concepts_qdhf}

Diversity maintenance in QD usually relies on manually designed diversity metrics to ensure a varied solution archive. However, such a requirement restricts the applicability of QD in more complex and open-ended domains, where the notion of diversity is likely to be abstract and qualitative. To address this, QD with automatic discovery of diversity has become popular. Instead of using pre-defined diversity metrics, recent work such as AURORA~\cite{cully2019autonomous,grillotti2022unsupervised} and RUDA~\citep{grillotti2022relevance} utilize unsupervised dimension reduction techniques to learn diversity metrics directly from the data.

One limitation of unsupervised methods is that the derived diversity metrics often capture the overall variance in \textit{current} solutions, which may not align well with the diversity needed for discovering novel and superior solutions. Such misalignment requires explicit mechanisms (\eg, RUDA) to guide the search, making the methods less applicable in complex and open-ended environments.
We introduce the new paradigm of QDHF, which offers greater flexibility compared to manually designing diversity metrics and outperforms unsupervised methods by leveraging diversity metrics that align to human intuition.

\section{Methods}
\label{sec:methods}

\begin{figure}[t]
  \vskip 0.1in
  \begin{center}
    \includegraphics[width=\linewidth]{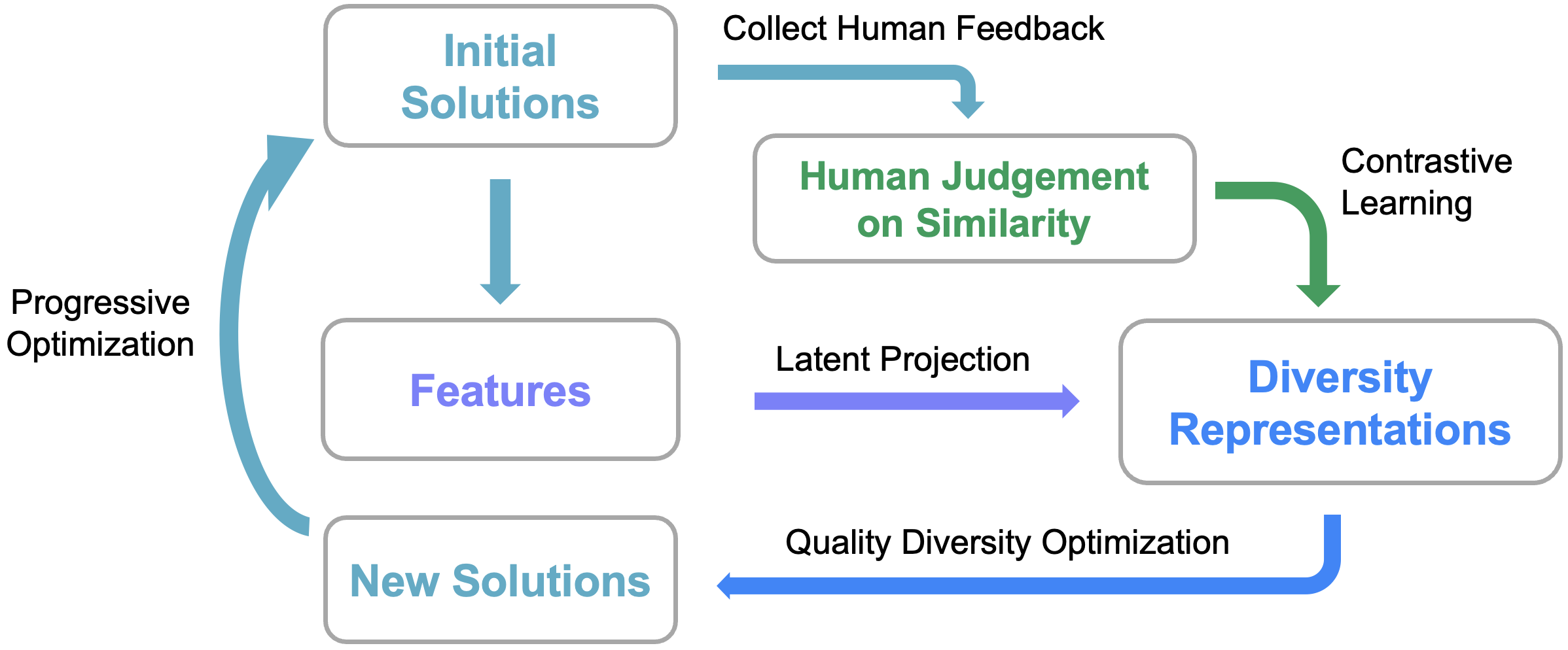}\\
    \vskip -0.05in
    \caption{Overview of the proposed QDHF method.}
    \label{fig:qdhf}
  \end{center}
  \vskip -0.2in
\end{figure}

This section first introduces a formulation of arbitrary diversity metrics from a generic learning perspective using latent projection, then describes how to align such metrics to human intuition of diversity using contrastive learning.
Finally, we introduce quality diversity through human feedback (QDHF), an online method for open-ended diversity-driven optimization. An overview is shown in Fig.~\ref{fig:qdhf}.

\subsection{Diversity Characterization with Latent Projection}
\label{sec:qdhf1}

Prior work~\citep{meyerson2016learning,cully2019autonomous,grillotti2022unsupervised} has shown that unsupervised dimensionality reduction algorithms such as principal component analysis (PCA) and auto-encoders (AE) can be utilized to automatically learn robot behavioral descriptors based on sensory data. In our framework, we first introduce a concept of \textit{diversity characterization}, which uses latent projection to transform arbitrary data into a semantically meaningful latent space that represents diversity. More specifically, given an input vector \(x\), we first employ a feature extractor \(f: \mathcal{X} \rightarrow \mathcal{Y}\) where \(\mathcal{X}\) denotes the input space and \(\mathcal{Y}\) the feature space. Post extraction, a projection function parameterized with $\theta$, denoted as \(D_r(y, \theta): \mathcal{Y} \rightarrow \mathcal{Z}\), is applied to project  \(y\) into a more compact latent representation:
\begin{equation}\label{eq:z}
  z = D_r(y, \theta), \text{where } y=f(x).
\end{equation}
\(\mathcal{Z}\) represents the latent space, wherein each dimension corresponds to a diversity metric.
The magnitude and direction capture various notions of diversity, offering a compact yet informative representation of \(x\).
For example, on a single dimension, smaller and larger values could indicate variations in certain characteristics such as the size of the object.
In this work, we use linear projection for simplicity and generality, although non-linear multi-layer projections are also compatible with our framework. The parameters for the projection are learned through a contrastive learning process described in the following section.

\subsection{Aligning Diversity Metrics to Human Intuition}

While unsupervised dimensionality reduction methods capture significant data variances, the resulting latents may not consistently offer useful semantics for optimization.
Recognizing this shortcoming, QDHF effectively aligns the latents to human notions of diversity through contrastive learning~\cite{chopra2005learning, schroff2015facenet}.

\paragraph{Contrastive learning.} Recent work has explored using contrastive learning as a type of self-supervised representation learning strategy~\cite{radford2021learning,chen2020simple,he2020momentum,tian2020contrastive}, and has demonstrated success of using it in modeling human preferences in RLHF~\citep{christiano2017deep} and image similarity~\citep{fu2023dreamsim}. Our framework takes a similar approach. Given a triplet of latent embeddings \(z_1\), \(z_2\), and \(z_3\), and supposing that the human input indicates that \(z_1\) is more similar to \(z_2\) rather than \(z_3\), our intention is to optimize the spatial relations of \(z_1\) and \(z_2\) in the latent space relative to \(z_3\). We use a triplet loss mechanism, \ie, to minimize the distance between \(z_1\) and \(z_2\) while maximizing between \(z_1\) and \(z_3\) via a hinge loss. This objective can be formalized as:
\begin{equation}\label{eq:l}
  \mathcal{L}(z_1, z_2, z_3) = \max(0, m + D(z_1, z_2) - D(z_1, z_3))
\end{equation}
where \(D(\cdot, \cdot)\) represents a distance metric in the embedding space, and \(m\) acts as a predetermined margin. Through this approach, the latent projections are aligned to both the inherent structure of data and human notions of similarity.

\paragraph{Human judgment on similarity.} To accommodate our design, we use Two Alternative Forced Choice (2AFC)~\citep{zhang2018unreasonable,fu2023dreamsim} to obtain human judgments on the similarity of solutions.
When presented with a triplet \(\{x_1, x_2, x_3\}\), an evaluator is prompted to discern whether \(x_2\) or \(x_3\) is more similar to the reference solution, \(x_1\).
Importantly, this mechanism works not only with human judgment but also with judgment produced by heuristics and AI systems, meaning that our framework remains universally applicable across different feedback modalities.

\subsection{Quality Diversity through Human Feedback}

We hereby introduce QDHF and its implementation through latent space projection and contrastive learning.

\paragraph{Objective.}
In QDHF, the diversity metrics $M_{\text{hf}, i}: \mathcal{X} \rightarrow \mathbb{R}$ are derived from human feedback on the similarity of solutions. Given a solution $x \in \mathcal{X}$, we define $M_{\text{hf}, i}(x) = z$ where $z\in \mathcal{Z}$ is the latent representation defined in Eq.~\ref{eq:z}. The latent space $\mathcal{Z}$ is used as the measurement space $M_{\text{hf}}(\mathcal{X})$, where each dimension $i$ in $\mathcal{Z}$ corresponds to a diversity metric $M_{\text{hf}, i}$. We can now specialize Eq.~\ref{eq:qd} for the objective of QDHF, $J_{\text{hf}}^*$, which is formulated as:
\begin{equation}\label{eq:qdhf2}
  J_{\text{hf}}^* = \sum_{i=1}^{s} \max_{x\in\mathcal{X}, z\in C_i} J(x)
\end{equation}
where $z = D_r(f(x), \theta)$ (Eq.~\ref{eq:z}). To effectively learn $\theta$, we use the contrastive learning mechanism (Eq.~\ref{eq:l}).

\paragraph{Training.}
One essential component of QDHF is the effective utilization of human feedback.
The initial diversity representations are learned from feedback collected on randomly generated solutions, which are often low-quality.
But as optimization proceeds, novel and higher-quality solutions are found, and the initial diversity metrics are unlikely to characterize them correctly.

We propose a progressive training strategy for QDHF where the latent projection is iteratively fine-tuned throughout the QD process.
As QD continually adds new solutions to its archive, we select samples from this evolving archive to gather human judgments on their similarity, producing online feedback used to fine-tune the latent projection.
Subsequently, the QD algorithm is re-initialized with current solutions but integrated with updated diversity representations.
This method ensures that the diversity representations remain relevant and reflective of the improved quality of solutions as optimization advances, supporting continuous and open-ended optimization for complex tasks.

\paragraph{Baseline method.}
We also propose a strong baseline method, `QDHF-Base,' which integrates the general design principles of RLHF and QD algorithms but learns diversity metrics offline. This method first collects human judgments on the similarity of randomly generated solutions, then learns the diversity representations (akin to reward model learning in RLHF), and finally uses these representations as diversity metrics in QD. QDHF-Base, although a baseline, is a novel algorithm for combining RLHF with QD optimization, leveraging the proven effectiveness of the RLHF framework.

\section{Theoretical Analysis}

In this section, we include further interpretations of QDHF, provide theoretical backing, and relate the advantages of QDHF to prior work in online learning.

\subsection{Learning Distinct Diversity Representations through Information Bottleneck}

When the dimensionality of the feature space, particularly in the last layer of a network, is significantly reduced, the process compels the network to encapsulate essential discriminative information within a limited set of features.
This process is effectively framed by the Information Bottleneck (IB) principle~\citep{tishby2000information}, which suggests that an optimal representation, \( Z \), of input data \( X \), should conserve only the information necessary for predicting the output \( Y \). The IB principle articulates this as an optimization task, aiming to balance the mutual information between \( Z \) and \( Y \) while constraining the information between \( X \) and \( Z \). Mathematically, this balance is captured by:
\begin{equation}
  \mathcal{L}_{IB} = I(X;Z) - \beta I(Z;Y).
\end{equation}
Here, a Lagrange multiplier \( \beta \) serves as a trade-off parameter, moderating the amount of information about \( X \) retained in \( Z \) and ensuring the relevance of \( Z \) for predicting \( Y \). More recent work~\citep{alemi2016deep} has further shown the IB principle’s pivotal role in deep neural networks' capacity to compress and selectively filter information.

In the context of contrastive learning, the objective is to optimize representations that carry the necessary information for predicting human preferences for similarity. Conventional approaches often employ large embedding dimensions to maximize the capability for similarity estimation, such as embeddings with 512 dimensions in DreamSim~\citep{fu2023dreamsim}.
In contrast, QDHF opts for significantly reduced dimensions to produce compact representations, imposing a constraint on the mutual information between \( Z \) and \( X \), formulated by:
\begin{equation}
  I(X;Z)\leq I_c
\end{equation}
where \(I_c\) is the information constraint. Intuitively, \(I_c\) forces \( Z \) to learn a minimal sufficient statistic of \( X \) for predicting \( Y \), resulting in \( Z \) being concise yet information-rich representations. The IB principle suggests that such representations should be both discriminative and succinct if \(\mathcal{L}_{IB}\) is minimal. Our empirical analysis, as described in Sec.~\ref{sec:abla}, further validates that when \(I(Z;Y)\) - the accuracy of predicting human judgments on the validation set - is high, QDHF can learn robust and informative diversity metrics and closely approaches the capability of QD using ground truth diversity.

Complementing this, the contrastive loss function (Eq.~\ref{eq:l}) minimizes the distance between similar pairs and maximizes it between dissimilar pairs, thereby encouraging the learning of meaningful representations for diversity that are aligned with human intuition. These representations can thus serve as robust diversity metrics in QD optimization, enhancing the model's generalization and creativity.

\subsection{Quality Diversity as an Active Sampling Strategy for Online Learning}

We also provide an analysis of QDHF from the perspective of online learning. Prior work has explored using online learning for similarity, especially in large-scale data environments.
QDHF introduces a nuanced methodology, enhancing traditional models like the Online Algorithm for Scalable Image Similarity (OASIS)~\citep{chechik2010large}. OASIS, part of the Passive-Aggressive (PA)~\citep{crammer2006online} algorithm family, optimizes similarity across image triplets \((p_i, p_i^+, p_i^-)\) to ensure a predefined margin in parameterized similarity scores \( S_W \), formalized as:
\begin{equation}
  l_W(p_i, p_i^+, p_i^-) = \max \left(0, 1 - S_W(p_i, p_i^+) + S_W(p_i, p_i^-)\right).
\end{equation}
This contrastive loss (equivalent to Eq.~\ref{eq:l} in the main paper by having a margin of $1$) leads to the global loss \(L_W\) defined as the sum of hinge losses over all triplets. The loss is minimized through iterative PA updates, where each iteration \(i\) involves optimizing \(W\) via:
\begin{equation}\label{eq:oasis}
  W_i = \underset{W}{\text{argmin}} \left( \frac{1}{2} \| W - W_{i-1} \|_F^2 + C\xi \right),
\end{equation}
subject to \( l_W(p_i, p_i^+, p_i^-) \leq \xi \) and \( \xi \geq 0 \), where \( \| \cdot \|_F \) denotes the Frobenius norm. The updates make sure that \( W_i \) optimizes the balance between staying close to the previous parameters \( W_{i-1} \) and reducing loss on the current triplet \( l_W(p_i, p_i^+, p_i^-) \), with \( C \) modulating this trade-off.

Conversely, QDHF focuses on QD as its core mechanism, utilizing contrastive learning to derive representations that serve as both benchmarks for diversity and guides for exploration. Unlike OASIS’s fixed-margin strategy, QDHF employs an adaptive scheme that refines its diversity metrics through continuous feedback and new data integration, fostering a progressing representation space that captures complex diversity-exploration interrelations. Mathematically, QDHF seeks to minimize the global loss as a blend of sustained learning from historical preference data and online data with feedback based on new triplet samplings:
\begin{equation}
  L_W = \sum_{\text{old triplets}} l_W + \sum_{\text{new triplets}} l_W,
\end{equation}
where \(\sum_{\text{old triplets}} l_W\) aims to retain the learned diversity, similar to how the first term \(\| W - W_{i-1} \|_F^2\) in Eq.~\ref{eq:oasis} ensures continuity. Theoretical analysis and empirical evidence from past studies~\citep{crammer2006online} suggest that such iterative online algorithms can achieve a cumulative online loss that remains small and converges over time.

While OASIS leverages a structured update rule to ensure that the learned similarity matrix progressively aligns with the target function, QDHF employs continuous feedback loops to develop a learning environment for diversity and exploration. This strategic feedback mechanism can be viewed as using QD as a sampling strategy for actively discovering unseen data for refining the learned diversity metrics, which ensures a vibrant, adaptive optimization landscape for online learning in more complex, open-ended environments.

\section{Experiments}  \label{sec:exp}

\subsection{Tasks and Benchmarks}
We describe our experimental setup across three benchmark tasks in the domains of robotics, RL, and generative modeling. For all experiments, we use MAP-Elites~\citep{mouret2015illuminating} as the QD algorithm for fair comparison. More implementation details can be found in Appendix~\ref{sec:imp}.

\paragraph{Robotic arm.}
We use the robotic arm domain from \citet{cully2015robots,vassiliades2018discovering}.
The goal is to identify an inverse kinematics solution for each accessible position of a planar robotic arm.
The objective function is to minimize the variance of the joint angles.
The standard way of measuring diversity is the endpoint's positions of the arm in the 2D space, which is computed using the forward kinematics of the arm, as detailed in \citet{murray2017mathematical}.

\paragraph{Maze navigation (RL).}
We adopt the Kheperax~\citep{grillotti2023kheperax} RL environment, which features a maze navigation task originally proposed in \citet{mouret2012encouraging}.
The goal is to discover a collection of neural network policy that controls the agent to navigate across varying positions in a maze. A Khepera-like robot is used as the agent, which is equipped with laser sensors positioned at the robot's facing directions for computing the distance. The maze is designed to be deceptive with immovable walls, making navigation a challenging optimization task.

\paragraph{Latent space illumination.}
The latent space illumination (LSI) task~\citep{fontaine2021illuminating} is designed for exploring the latent space of a generative model.
LSI initially aims to generate different gameplay levels and has later been extended to generate images of human faces that correspond to specific text prompts as diversity~\citep{fontaine2021differentiable}. We introduce a new LSI benchmark on text-to-image generation with Stable Diffusion~\citep{rombach2022high}, a latent diffusion model with state-of-the-art capability. In this task, the goal is to find high-quality (to match a text prompt, scored by CLIP~\citep{radford2021learning}) and diverse images by optimizing the latent vectors, where the diffusion model is treated as a black box. There are no pre-defined diversity metrics, making this task a more challenging and open-ended optimization problem in a real-world setting.

This benchmark also features two types of text prompts as subtasks: singular and compositional prompts. Singular prompts elicit responses about a single concept, usually taking the form of ``an image of a \textit{category}'' where \textit{category} is drawn from labels in popular image datasets such as COCO~\citep{lin2014microsoft} and ImageNet~\citep{russakovsky2015imagenet}. We also include a few popular text prompts used in prior work~\citep{rombach2022high} that have a similar form. For singular prompts, diversity gives more variation in images that have the same semantics, which can better meet the diverse preferences of users.

Compositional prompts are complex conjunctions of multiple concepts. We use the MCC-250 benchmark~\citep{bellagente2023multifusion}, which consists of prompts describing two objects with respective attributes, \eg, ``a red apple and a yellow banana''. Recent work has shown that diffusion models often fail to correctly compose the specific objects and attributes~\citep{chefer2023attend,bellagente2023multifusion}, but enhanced diversity may explore different ways of composing, and thus improve the chance of generating the correct image given a limited budget of responses.

\begin{table*}[t]
  \caption{Results for robotic arm. We report the QD score (normalized to a scale of 0-100) and coverage for ``All Solutions'' (solutions found throughout training) and ``Archive Solutions'' (solutions in the final archive). QDHF significantly outperforms QDHF-Base and AURORA, and closely approaches the search capability of QD using ground truth diversity metrics when considering all solutions.}
  \label{tb:arm}
  \vskip 0.1in
  \begin{center}
    \begin{small}
      \begin{tabular}{lcccc}
        \toprule
                         & \multicolumn{2}{c}{All Solutions} & \multicolumn{2}{c}{Archive Solutions}                                                     \\
        \cmidrule(lr){2-3}\cmidrule(lr){4-5}
        Methods          & QD Score                          & Coverage                              & QD Score                & Coverage                \\
        \midrule
        AURORA-Pre (AE)  & $38.5 \pm 12.7$                   & $56.2 \pm 6.8$                        & $14.3 \pm 6.3$          & $20.3 \pm 6.4$          \\
        AURORA-Inc (AE)  & $53.0 \pm 9.4$                    & $63.2 \pm 6.2$                        & $17.6 \pm 4.1$          & $18.9 \pm 4.1$          \\
        AURORA-Pre (PCA) & $38.4 \pm 13.6$                   & $54.1 \pm 9.0$                        & $14.2 \pm 4.4$          & $19.3 \pm 5.5$          \\
        AURORA-Inc (PCA) & $45.9 \pm 6.4$                    & $59.0 \pm 3.7$                        & $18.3 \pm 3.4$          & $19.0 \pm 3.5$          \\
        \midrule
        QDHF-Base        & $54.5 \pm 4.3$                    & $62.7 \pm 2.7$                        & $31.8 \pm 4.5$          & $34.1 \pm 4.2$          \\
        QDHF             & $\mathbf{72.5 \pm 0.9}$           & $\mathbf{77.3 \pm 1.2}$               & $\mathbf{56.4 \pm 0.9}$ & $\mathbf{59.9 \pm 0.9}$ \\
        \midrule
        QD-GT            & $74.8 \pm 0.2$                    & $79.5 \pm 0.3$                        & $74.8 \pm 0.2$          & $79.5 \pm 0.3$          \\
        \bottomrule
      \end{tabular}
    \end{small}
  \end{center}
  \vskip -0.2in
\end{table*}

\subsection{Experimental Design and Evaluation}\label{sec:design}

In our experiments, we design two distinct scenarios: structured tasks, where ground truth diversity metrics are available, and open-ended tasks, where such metrics are not.

\paragraph{Structured tasks.}
The first scenario leverages a predefined ground truth diversity metric to simulate human feedback.
The primary reason for using simulated feedback is to facilitate evaluation and comparisons, which requires measuring diversity consistently across different methods.
The robotic arm and maze navigation tasks fall under this category, where the ground truth diversity metrics correspond to the position (x and y values) of the arm and of the robot in the 2D space, respectively. The ``human judgment'' is determined by the similarity of the positions, calculated as the L2 distance from the ground truth measurements.

To validate the effectiveness of QDHF, we benchmark against AURORA~\citep{grillotti2022unsupervised,cully2019autonomous} and standard QD. The standard QD uses ground truth diversity metrics, which offers an oracle control representing the best possible performance of QD.
For comprehensiveness, we implement four variants of AURORA, encompassing two dimension-reduction techniques: PCA and AE, and two training strategies:  pre-trained (Pre) and incremental (Inc).

For evaluation, solutions of each method are additionally stored in a separate archive corresponding to the ground truth diversity metrics.
We report QD score~\citep{pugh2015confronting} (sum of objective values, Eq.~\ref{eq:qd}) and coverage (ratio of filled cells to total cells), which are standard metrics for assessing both quality and diversity of solutions. The evaluation is conducted in two settings: 1) solutions in the final archive, and 2) solutions discovered by the algorithm throughout its entire search process. The first setting evaluates the alignment of learned diversity metrics with the ground truth metrics, and the second setting provides insights into the overall efficacy of the search process regardless of how the diversity is measured and maintained.

\paragraph{Open-ended tasks.} In the second scenario, there is no ground truth diversity metric and real human feedback data is used in training and evaluation, which applies to the LSI task. To facilitate the efficiency and scalability of our method, instead of having human labelers in the loop, we train a preference model with real human feedback data and use the preference model to source feedback for training and evaluation. Similar approaches have been widely used in recent RLHF applications~\citep{stiennon2020learning,ouyang2022training}.

We use human judgment data from the NIGHTS dataset, and the DreamSim model to estimate image similarity, both from \citet{fu2023dreamsim}. For comparisons, we implement a best-of-n approach on Stable Diffusion as the baseline, which generates the same number of images as QDHF with random latents, and select a solution set with top CLIP scores. We also propose a stronger baseline (Best-of-n+) featuring a heuristic that increases the diversity of the latents by sampling the next latent to be distant from previous ones.

The LSI experiments are conducted using 100 singular and 100 compositional prompts.
Quantitatively, we report the average CLIP score for assessing how well the generated images match the prompt. We also use DreamSim to calculate the pairwise distance between images in the solution to measure diversity. The mean pairwise distance indicates the average separation of images, and the standard deviation indicates the variability in how the images are distributed. We also conduct human evaluation ($n=50$) using Amazon Mechanical Turk to qualitatively evaluate the results, and show examples of solutions for qualitative assessment.

\begin{table*}[t]
  \caption{Results for maze navigation. QDHF shows superior performance compared to QDHF-Base and AURORA, as well as a resemblance of the search capability of QD with ground-truth diversity.}
  \label{tb:maze}
  \vskip 0.1in
  \begin{center}
    \begin{small}
      \begin{tabular}{lcccc}
        \toprule
                         & \multicolumn{2}{c}{All Solutions} & \multicolumn{2}{c}{Archive Solutions}                                                     \\
        \cmidrule(lr){2-3}\cmidrule(lr){4-5}
        Methods          & QD score                          & Coverage                              & QD score                & Coverage                \\
        \midrule
        AURORA-Pre (AE)  & $35.9 \pm 0.6$                    & $38.1 \pm 0.9$                        & $22.3 \pm 0.6$          & $23.0 \pm 0.8$          \\
        AURORA-Inc (AE)  & $40.0 \pm 2.1$                    & $46.7 \pm 4.9$                        & $19.4 \pm 1.2$          & $22.8 \pm 2.0$          \\
        AURORA-Pre (PCA) & $35.5 \pm 0.4$                    & $37.8 \pm 0.4$                        & $22.9 \pm 0.6$          & $23.7 \pm 0.7$          \\
        AURORA-Inc (PCA) & $39.0 \pm 0.8$                    & $45.3 \pm 3.5$                        & $18.0 \pm 0.7$          & $21.0 \pm 1.1$          \\
        \midrule
        QDHF-Base        & $35.6 \pm 0.6$                    & $37.9 \pm 1.1$                        & $\mathbf{23.7 \pm 0.9}$ & $24.4 \pm 1.1$          \\
        QDHF             & $\mathbf{42.0 \pm 1.7}$           & $\mathbf{51.3 \pm 5.5}$               & $22.5 \pm 1.3$          & $\mathbf{27.2 \pm 3.0}$ \\
        \midrule
        QD-GT            & $42.7 \pm 2.7$                    & $52.7 \pm 7.1$                        & $42.7 \pm 2.7$          & $52.6 \pm 7.0$          \\
        \bottomrule
      \end{tabular}
    \end{small}
  \end{center}
  \vskip -0.2in
\end{table*}

\begin{table}[t]
  \caption{Quantitative Results for LSI (singular prompts). We report the CLIP score and DreamSim pairwise distance (mean and std.) as quantitative metrics. QDHF demonstrates competitive quality to Stable Diffusion (Best-of-n) measured by CLIP score, and significantly better diversity measured by pariwise distance.}
  \label{tab:lsi_quan}
  \vskip 0.1in
  \begin{center}
    \begin{small}
      \begin{tabular}{lccc}
        \toprule
        Method          & CLIP Score       & Mean PD          & Std. PD          \\
        \midrule
        SD (Best-of-n)  & $67.64$          & $0.405$          & $0.98$           \\
        SD (Best-of-n+) & $67.87$          & $0.412$          & $0.102$          \\
        QDHF            & $\mathbf{68.47}$ & $\mathbf{0.517}$ & $\mathbf{0.153}$ \\
        \bottomrule
      \end{tabular}
    \end{small}
  \end{center}
  \vskip -0.2in
\end{table}

\begin{table*}[t]
  \caption{Human evaluation results for LSI. In a blind user experience study, QDHF outperforms Stable Diffusion (best-of-n) by considerable margins in terms of user preference and perceived diversity for singular prompts and shows a clear advantage in correctness by human judgment for compositional prompts.}
  \label{tab:lsi_qual}
  \vskip 0.1in
  \begin{center}
    \begin{small}
      \begin{tabular}{lccc}
        \toprule
                                  & \multicolumn{2}{c}{Singular Prompts} & \multicolumn{1}{c}{Compositional Prompts}                           \\
        \cmidrule(lr){2-3}\cmidrule(lr){4-4}
        Method                    & User Preference                      & Perceived Diversity                       & Judgment on Correctness \\
        \midrule
        \textit{Cannot Determine} & $9.7\%$                              & $8.6\%$                                   & $6.4\%$                 \\
        SD (Best-of-n)            & $31.9\%$                             & $24.8\%$                                  & $40.6\%$                \\
        QDHF                      & $\mathbf{58.4\%}$                    & $\mathbf{66.7\%}$                         & $\mathbf{52.8\%}$       \\
        \bottomrule
      \end{tabular}
    \end{small}
  \end{center}
  \vskip -0.2in
\end{table*}

\begin{figure*}[t]
  \vskip 0.1in
  \begin{center}
    \includegraphics[width=\linewidth]{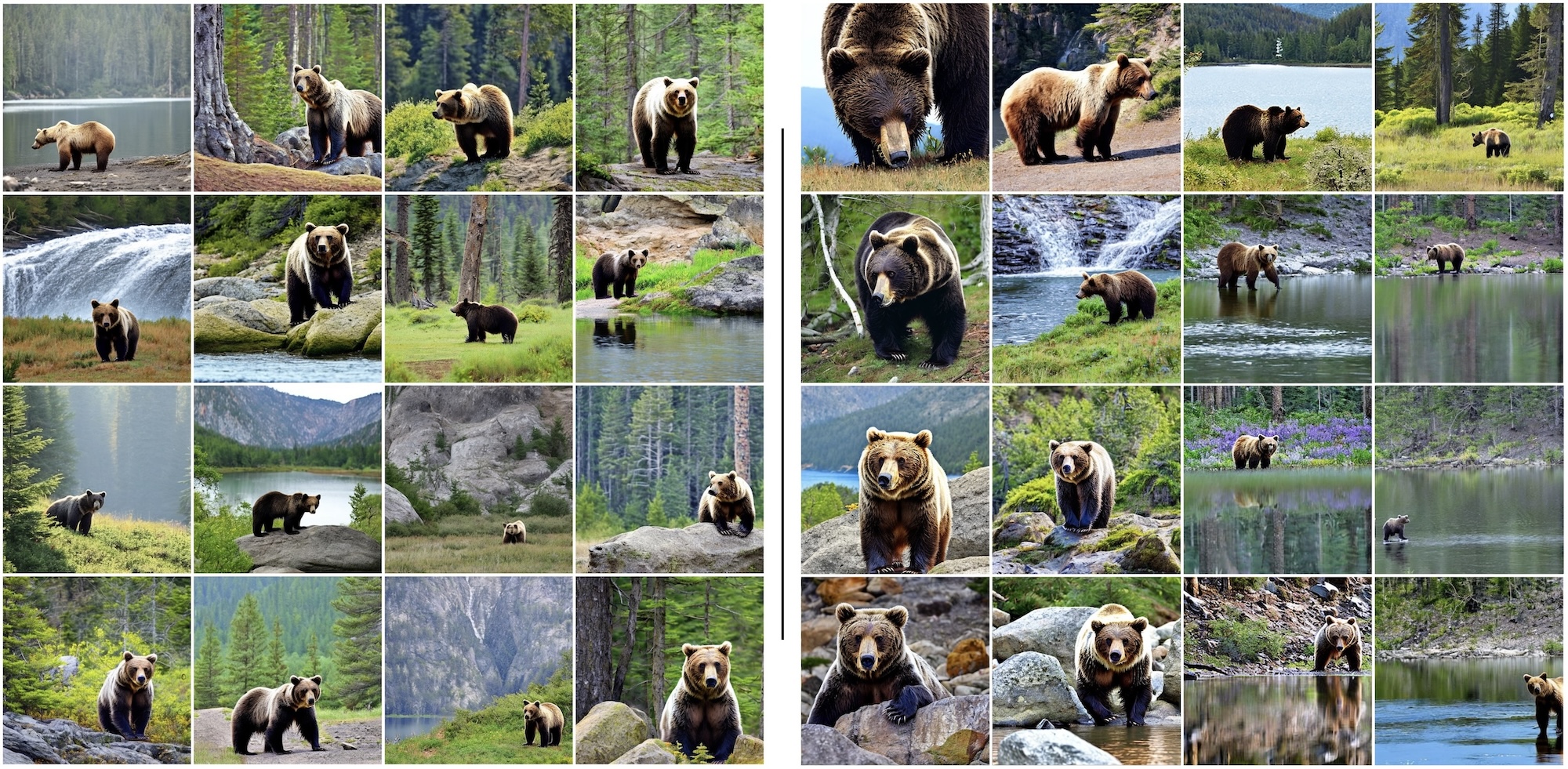}\\
    \vskip -0.05in
    \caption{Qualitative result for LSI (singular prompts). The target prompt is ``an image of a bear in a national park''. The left 4x4 grid displays Stable Diffusion (Best-of-n) results, \ie, randomly generated images with the highest CLIP scores. The right grid displays a uniformly sampled subset of QDHF solutions. Qualitatively, images generated by QDHF have more variations and show visible trends of diversity such as object sizes (large to small along the x-axis) and landscape types (rocky to verdant along the y-axis, terrestrial to aquatic along the x-axis). The selected example represents the average user preference ratio observed in our user experience study, where QDHF results are about twice as preferred and three times considered as more diverse.}
    \label{fig:lsi}
  \end{center}
  \vskip -0.2in
\end{figure*}

\begin{figure}[t]
  \vskip 0.1in
  \begin{center}
    \includegraphics[width=\linewidth]{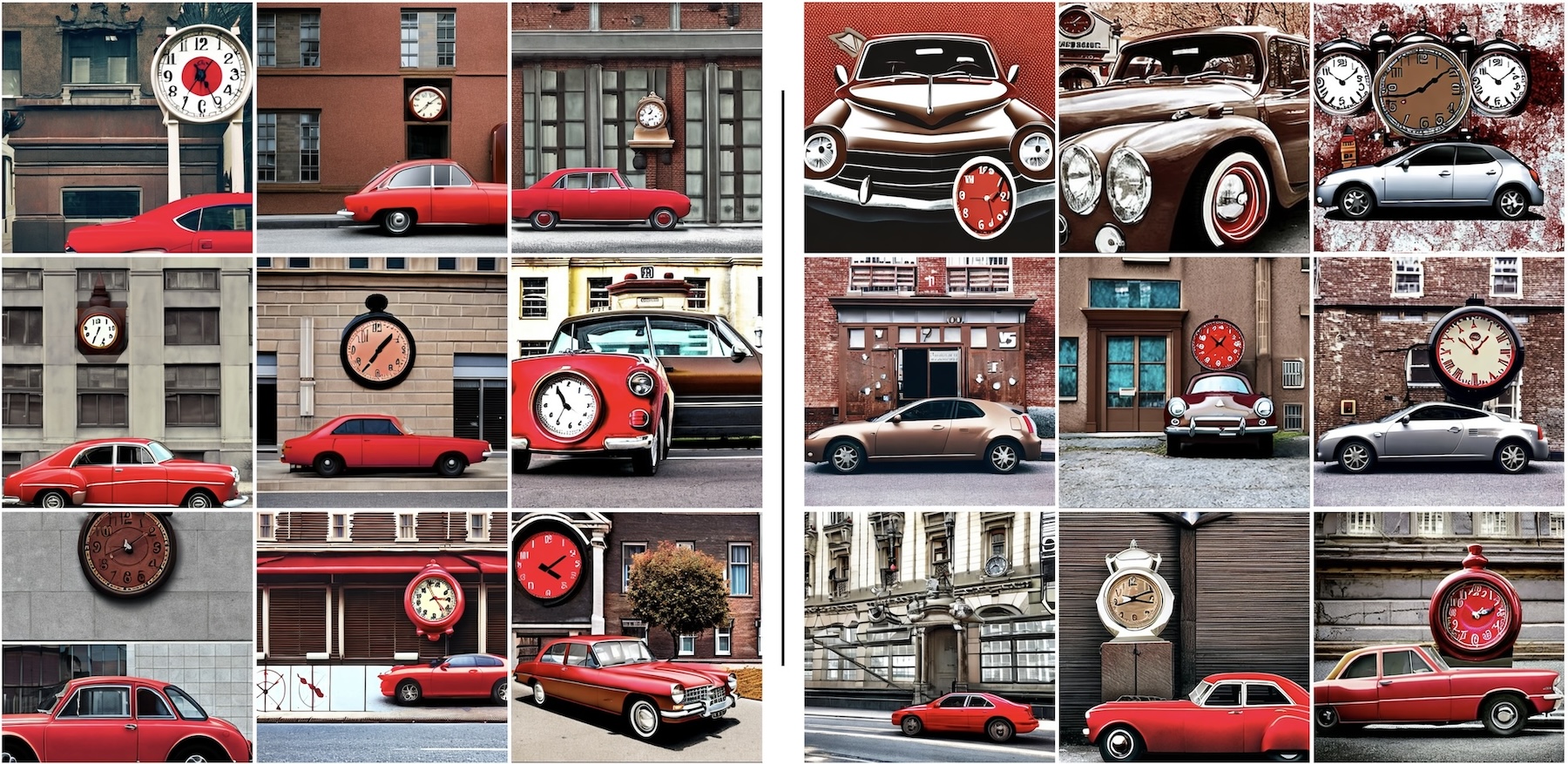}\\
    \vskip -0.05in
    \caption{Qualitative result for LSI (compositional prompts). The target prompt is ``a \textcolor{brown}{brown} car and a \textcolor{red}{red} clock''. The left 3x3 grid displays SD (Best-of-n) results, where we can observe the issue of attribute leakage, \ie, the color of the clock (\textcolor{red}{red}) leaks to the car. However, QDHF (on the right) can generate conceptually correct responses by having more variation in the responses.}
    \label{fig:lsi_comp}
  \end{center}
  \vskip -0.2in
\end{figure}

\subsection{Results}

We present our major results and findings below, with more detailed results available in Appendix~\ref{sec:aresults}.

\paragraph{Robotic arm.}

For the robotic arm task, detailed results are presented in Table~\ref{tb:arm}. The statistics are accumulated over 20 repeated trials. QDHF significantly surpasses AURORA in terms of both QD score and coverage. The results highlight QDHF's capability to enhance QD algorithms over unsupervised diversity discovery methods.
QDHF also outperforms QDHF-Base, a variant that separates the processes of diversity learning and QD optimization. Such results validate our hypothesis that diversity representations need to be refined during optimization, as finding more high-quality solutions changes the distribution of the solutions in the search space.
It is worth highlighting that QDHF closely matches the performance of QD with the ground truth metric when evaluating all solutions, which indicates that QDHF is a competitive alternative for QD in situations where manually designed metrics are not available.

\paragraph{Maze navigation.}

Results are shown in Table~\ref{tb:maze}. The statistics are accumulated over 10 repeated trials since maze navigation is more computationally expensive. Similar to the robotic arm task, we observe that QDHF surpasses QDHF-Base and AURORA, and closely matches the performance of standard QD. The results further support the validity of QDHF as an alternative to standard QD with improved flexibility and competitive performance.

\paragraph{Latent space illumination.}

Results are shown in Table~\ref{tab:lsi_quan}. The results are summarized over 100 text prompts, and we include more details in Appendix~\ref{sec:alsi}. SD (Best-of-n) is the best-of-n responses produced by Stable Diffusion, and SD (Best-of-n+) is the best-of-n with a heuristic to enhance diversity in sampling (Sec. ~\ref{sec:design}). Quantitatively, QDHF has a similar CLIP score to both baseline methods, but much higher mean and standard deviation of pairwise distance, which indicates that QDHF is able to generate more diverse solutions while maintaining the high-quality. We also observe that SD (Best-of-n+) does not produce better diversity, which indicates that there exists a strong inductive bias in the diffusion model and that generating diverse images is a challenging latent optimization problem.

We also compare QDHF and SD (Best-of-n) with human evaluation in a blind user study.
For singular prompts, QDHF outperforms SD (Best-of-n) with a considerable margin on both the user preference ratio and user diversity perception ratio. Notably, we find that most users think QDHF generates more diverse images, and a majority of them also prefer to have such diversity in the solution. An example of the LSI results is depicted in Fig.~\ref{fig:lsi}, where QDHF results have more variations and show visible trends of diversity.

For compositional prompts, users are asked to select which 3x3 set of generated images most accurately responds to the prompt. We find that QDHF is more accurate in generating images with correct concepts as evaluated by humans. Fig.~\ref{fig:lsi_comp} shows an example, where QDHF  mitigates the common attribute leakage issue in compositional image generation by improving the diversity of responses. It is worth highlighting that QDHF serves as a flexible black-box optimization method, compatible with any existing generative models, such as Stable Diffusion, without requiring access to or fine-tuning the model parameters.

\begin{figure}[t]
  \vskip 0.1in
  \begin{center}
    \includegraphics[width=.49\linewidth]{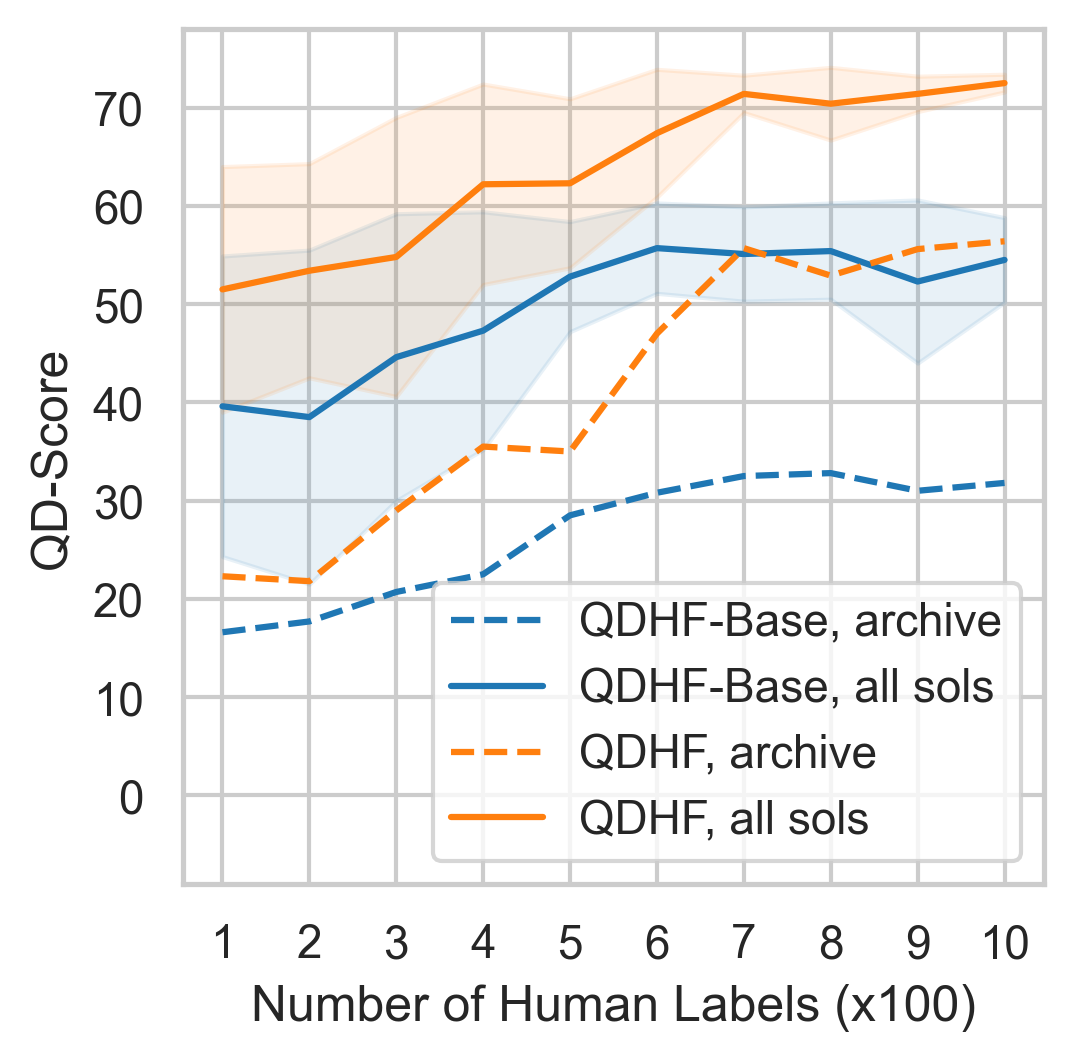}
    \includegraphics[width=.5\linewidth]{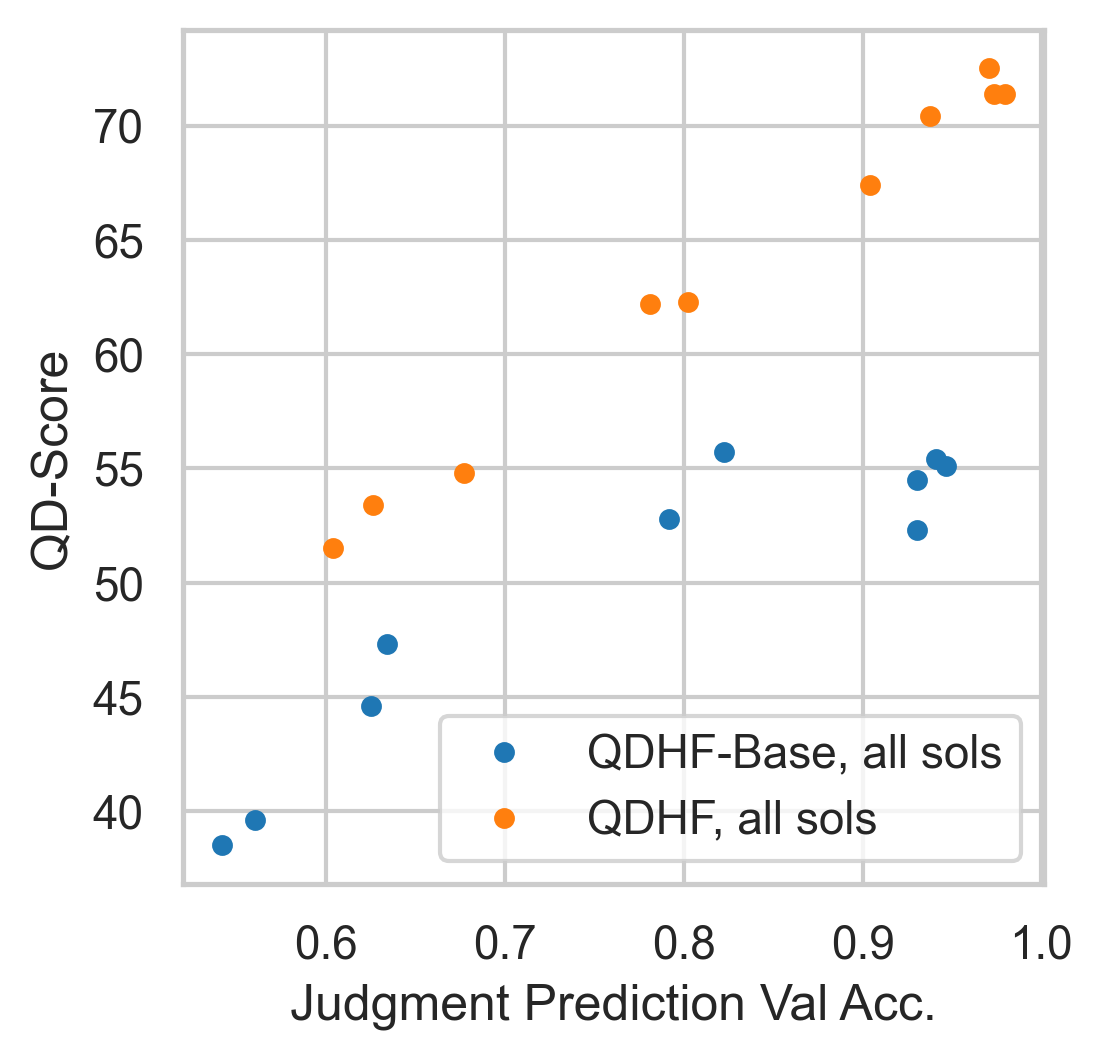}\\
    \vskip -0.05in
    \caption{Analysis of varying human feedback sample sizes on robotic arm. ``Judgment prediction val acc'' is the accuracy of the latent projection in predicting human preferences based on a validation set. There is a direct correlation between QD score and sample size, with QDHF's performance closely tied to the accuracy of latent projection in reflecting human judgment.}
  \end{center}
  \vskip -0.2in
  \label{fig:sample}
\end{figure}

\subsection{Ablation Studies and Analysis}\label{sec:abla}

We highlight the analysis of QDHF's scalability and quality of learned diversity metrics. Additional analysis on robustness is provided in Appendix~\ref{sec:arobotic}.

\paragraph{Scalability.}
To ensure scalability, determining the necessary amount of human feedback to achieve sufficient performance is critical. To investigate this, we perform a study of varying sample sizes of human judgments, shown in Fig.~\ref{fig:sample}. The left plot demonstrates the relationship between QD score and sample size, showing a strong correlation. Diving deeper into this relationship, we evaluate how accurately the latent projection captures the diversity and mirrors human judgment using the accuracy of predicting human judgment on a validation set. The right plot reveals a strong correlation between accuracy and QD score, which suggests that QDHF relies on learning useful diversity metrics. In other words, by evaluating the accuracy of judgment prediction during training, we can estimate whether current feedback data is sufficient for QDHF to perform well.

Secondly, we note that the judgment data for QDHF does not need to come entirely from humans. In the LSI experiments, we use a preference model (DreamSim) to source human feedback during training QDHF, and we show that an accurate preference model can be used for QDHF as an alternative to human labelers. Combining these two observations, we conclude that QDHF has good scalability towards more complex tasks because 1) we can anticipate its performance through online validation, and 2) the samples used by QDHF can be sourced from a preference model trained on a fixed amount of human labor.

\begin{figure}[t]
  \vskip 0.1in
  \begin{center}
    \includegraphics[width=\linewidth]{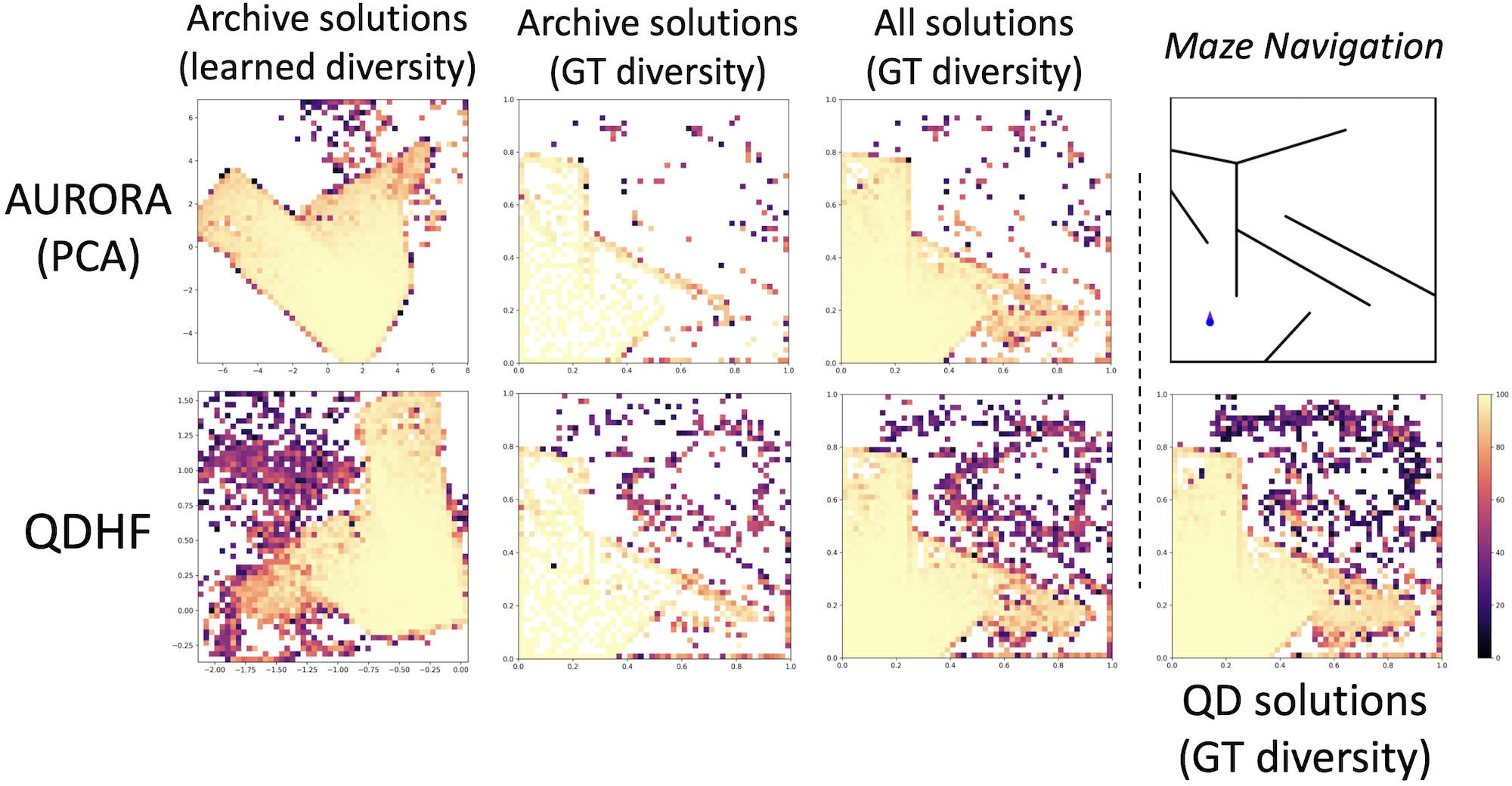}\\
    \vskip -0.05in
    \caption{Visualization of maze navigation solutions in different diversity spaces. ``GT'' stands for ground truth. Each point on the heatmap is a solution with its objective value visualized in color. QDHF fills up the archives with more solutions than AURORA. While both methods learned a rotated version of the maze as the diversity metrics (first column), QDHF more accurately captures the scale of the maze, especially in under-explored areas.}
    \label{fig:archive}
  \end{center}
  \vskip -0.2in
\end{figure}

\paragraph{Alignment between learned and ground truth diversity.}
We evaluate the alignment between diversity metrics derived by QDHF and the underlying ground truth diversity metrics. In Fig.~\ref{fig:archive}, the archives of QDHF and AURORA-Inc (PCA) are visualized for maze navigation. Both AURORA and QDHF appear to effectively learn a diversity space reflective of the ground truth. However, QDHF exhibits an enhanced capability to discern the relative distances between solutions, especially in under-explored areas. This indicates QDHF's strength in identifying novel solutions, and this efficacy stems from QDHF's ability to better align its learned diversity space to the ground truth diversity, especially concerning the scales. A similar analysis for the robotic arm task is included in Appendix~\ref{sec:arobotic}, where QDHF also shows its advantage in modeling the scales of diversity representations for enhanced discovery of novel solutions.

\section{Related Work}

\paragraph{Learning from human feedback.}
This work expands upon recent developments in methodologies for aligning models to human objectives, especially reinforcement learning from human feedback (RLHF)~\citep{christiano2017deep,ibarz2018reward}. RLHF was initially proposed to train RL agents in simulated environments such as Atari games. It has since been applied to fine-tune or perform one-shot learning on language models for tasks including text summarization~\citep{ziegler2019fine,stiennon2020learning,wu2021recursively}, dialogue~\citep{jaques2019way}, and question-answering~\citep{nakano2021webgpt,bai2022training,ouyang2022training}, as well as vision tasks such as measuring perceptual similarity~\citep{fu2023dreamsim}. While past efforts have focused on learning reward or preference models from human intentions, we propose to learn diversity metrics through human feedback, which then drives the optimization process in QD algorithms.

\paragraph{Diversity-driven optimization.}
Instead of optimizing for a single optimal solution, diversity-driven optimization methods such as QD~\citep{lehman2011evolving,cully2015robots,mouret2015illuminating, pugh2016quality} and Novelty Search~\citep{lehman2011abandoning} aim to identify a variety of (top-performing) solutions that exhibit novelty or diversity. Prior work expands on QD by enhancing diversity maintenance~\citep{fontaine2019mapping,smith2016rapid,vassiliades2017using}, the search process~\citep{fontaine2020covariance,vassiliades2018discovering,nordmoen2018dynamic,sfikas2021monte}, optimization mechanism~\citep{kent2020bop,conti2018improving,fontaine2021differentiable}, and exploring implicit methods for diversity-driven optimization~\citep{ding2022optimizing,boldi2023objectives,ding2023probabilistic,ni2024dalex,spector2024particularity}.
Recent work~\citep{grillotti2022relevance,grillotti2022unsupervised,cully2019autonomous} also explores unsupervised methods for diversity discovery. Our work differs by leveraging human feedback to derive diversity metrics that are aligned with human interest, and thus often more beneficial for optimization.

\section{Conclusion}
This paper introduces Quality Diversity through Human Feedback (QDHF), which expands the reach of QD by leveraging human feedback to effectively derive diversity metrics.
Empirical results show that QDHF outperforms current unsupervised diversity discovery methods, and compares well to standard QD that uses manually crafted diversity metrics.
In particular, applying QDHF in an open-ended generative task substantially enhances the diversity of text-to-image generations.
We provide an analysis of QDHF's scalability, robustness, and the quality of diversity metrics learned.
Future work will focus on applying QDHF to more challenging tasks in complex and open-ended environments.

\section*{Acknowledgments}
Li Ding and Lee Spector were supported by the National Science Foundation under Grant No. 2117377.
Jeff Clune and Jenny Zhang were supported by the Vector Institute, a grant from Schmidt Futures, an NSERC Discovery Grant, and a generous donation from Rafael Cosman. Any opinions, findings, and conclusions or recommendations expressed in this publication are those of the authors and do not necessarily reflect the views of the funding agencies.
The authors would like to thank Andrew Dai and Herbie Bradley for insightful suggestions, Bryon Tjanaka for help preparing the tutorial, members of the PUSH lab at Amherst College, University of Massachusetts Amherst, and Hampshire College for helpful discussions, and anonymous reviewers for their thoughtful feedback.
This work utilized resources from Unity, a collaborative, multi-institutional high-performance computing cluster at the Massachusetts Green High Performance Computing Center (MGHPCC).

\section*{Impact Statement}
This paper presents work whose goal is to advance the field of Machine Learning. There are many potential societal consequences of our work, none of which we feel must be specifically highlighted here.

\bibliography{qdhf}
\bibliographystyle{icml2024}

\newpage
\appendix

\input{appendix.tex}  

\end{document}

%% file: appendix.tex
\section{Preliminaries on Quality Diversity} \label{sec:apqd}

Quality Diversity (QD)~\citep{mouret2015illuminating,cully2015robots,pugh2016quality,lehman2011evolving} is a concept in the field of optimization and artificial intelligence that emphasizes not just finding the best possible solution to a problem (quality), but also discovering a variety of good solutions that are diverse in their characteristics (diversity). This approach is particularly valuable in complex problem-solving scenarios where there might be multiple good solutions, each with its unique benefits.

\subsection{Quality Diversity}

The QD process aims to find the best representative samples, not only seeking the absolute best but also ensuring that the selections are varied and uniquely excellent in their own ways.
Intuitively, imagine assembling a soccer team with QD: it meticulously recruits top-tier players across various positions to build a well-rounded team, rather than simply gathering the most renowned players regardless of their specialized roles. Key aspects of QD include:

\begin{itemize}[topsep=0pt,parsep=0pt]
  \item Quality: This refers to how well a solution meets the desired criteria or objectives. In QD, the aim is to identify solutions that are highly effective or optimal with respect to the goals of the task.
  \item Diversity: Unlike traditional optimization that focuses on the single best solution, QD seeks a range of good solutions that are different from each other. This diversity can be in terms of features, approaches, or strategies the solutions employ.
  \item Exploration and Exploitation: QD balances exploration (searching for new, diverse solutions) and exploitation (refining known good solutions). This balance helps navigate the solution space and uncover unique solutions that may be overlooked by conventional methods.
\end{itemize}

\subsection{MAP-Elites}

\begin{algorithm}[tb]
  \caption{MAP-Elites Algorithm}
  \label{alg:map}
  \begin{algorithmic}[1]
    \STATE Initialize a map of solutions, each cell representing a unique feature combination
    \WHILE{not converged}
    \STATE Generate new solutions via mutation and crossover
    \FOR{each solution}
    \STATE Evaluate the solution for its performance and feature characteristics
    \STATE Identify the corresponding cell in the map based on features
    \IF{solution is better than the current cell occupant}
    \STATE Replace the cell's solution with the new solution
    \ENDIF
    \ENDFOR
    \ENDWHILE
    \STATE Return the map of elite solutions
  \end{algorithmic}
\end{algorithm}

MAP-Elites~\citep{mouret2015illuminating} stands out in evolutionary computation for its unique approach to exploring solution spaces. Unlike traditional algorithms that target a single optimal solution, MAP-Elites focuses on revealing a broad spectrum of high-performing solutions, categorized by distinct features. A high-level view of MAP-Elites, outlining its core steps, is shown in Algorithm~\ref{alg:map}.

The algorithm uses a grid or map where each cell corresponds to a unique combination of feature descriptors. New solutions are generated through mutation, and are evaluated for their performance and feature characteristics. The map is updated continually, with each cell holding the best-performing solution for its feature combination, ensuring a rich diversity of high-quality solutions.

MAP-Elites is particularly advantageous in domains requiring adaptability and robustness, such as robotics, or in areas where creativity and a wide range of solutions are beneficial, like design and art. It provides insights into the solution space, highlighting the relationship between different solution features and their trade-offs.

\section{Additional Implementation Details}
\label{sec:imp}

In this section, we detail our implementation and hyperparameters used in the experiments.

\paragraph{For all tasks.}

The frequency of diversity representation fine-tuning in QDHF decreases exponentially over time as the learned metrics become more robust. In our experiments, the latent projection is updated 4 times at iteration $1$, $10\%\cdot n$, $25\%\cdot n$, and $50\%\cdot n$ for a total of $n$ iterations. To fairly compare with QDHF-base, each update consumes 1/4 of the total budget of human feedback.

\paragraph{Robotic arm.}

The robotic arm repertoire task is configured to have 10 degrees of freedom, \ie, the solution is a vector of 10 values, each specifying a joint angle. For QDHF and AURORA, we extract the features from the raw solution by running the accumulated sum on the solution vector and applying $\sin$ and $\cos$ transformations, resulting in a 20-dim feature vector. The latent projection in QDHF transforms the feature into a 2-dim embedding. For AURORA, the auto-encoder has an architecture of 64-32-2-32-64 neurons in each layer, where the mid-layer is the embedding used for QD. For QDHF, we use 1,000 judgments of simulated human feedback. The ground truth diversity is given by the end-point of the arm. For all experiments, we run MAP-Elites for 1000 iterations, and for each iteration, we generate a batch of 100 solutions with Gaussian mutation (adding Gaussian noises sampled from $\mathcal{N}(0, 0.1^2)$), and evaluate them. The archive has a shape of $(50,50)$, \ie, each of the 2 dimensions is discretized into 50 equal-sized bins.

\paragraph{Maze navigation.}

For the maze navigation task, the robot features two contact sensors, and each of these sensors yields a value of 1 upon contact with a wall and -1 in the absence of any contact.
The robot operates within a bounded square maze comprised of immovable, straight walls, which prevents the robot from moving upon collision. The episode length of the environment is 250. The solution is the network parameters of the default MLP policy network with a hidden-layer size of 8. We evaluate the policy and obtain the state descriptors. The objective is the accumulated reward at each state. For diversity measures, the ground truth diversity is the end-position of the agent, \ie, the position at the last state. For QDHF and AURORA, we extract features from the state descriptor as the x and y positions of the agent at each state. The latent projection in QDHF transforms the feature into a 2-dim embedding. For AURORA, the auto-encoder has the same architecture of 64-32-2-32-64 nodes in each layer, where the mid-layer is the embedding used for QD. For QDHF, we use 200 judgments of simulated human feedback. For all experiments, we run MAP-Elites for 1000 iterations, and for each iteration, we generate a batch of 200 solutions with Gaussian mutation (adding Gaussian noises sampled from $\mathcal{N}(0, 0.2^2)$), and evaluate them. The archive has a shape of $(50,50)$.

\paragraph{Latent space illumination.} In the LSI task, we use human judgment data from the NIGHTS dataset, and the DreamSim model to estimate image similarity, both from \citet{fu2023dreamsim}. For comparisons, we implement a best-of-n approach on Stable Diffusion as the baseline, which generates a total of 2,000 images (the same number generated by QDHF) with latents randomly sampled from uniform $\mathcal{U}(0,1)$, and select a solution set of 400 images (the same number of solutions to be produced by QDHF) with the top CLIP scores. We also propose a stronger baseline featuring a heuristic (Best-of-n+), which works as follows: Starting from the second image to be generated, we sample 100 latents randomly and choose the one with a maximal l2 distance to the previously sampled latent. This approach increases the diversity of the latents and could potentially also increase the diversity of responses.

We run QDHF for 200 iterations with a batch size of 5 solutions per iteration.
The solutions are generated with Gaussian mutation (adding Gaussian noises sampled from $\mathcal{N}(0, 0.1^2)$). The archive has a shape of $(20, 20)$. The solution is the latent vector used as the input to Stable Diffusion, which has a shape of $(4, 64, 64)$. We use Stable Diffusion v2.1-base, which generates images at a resolution of 512x512. The feature extractor is a CLIP model with ViT-B/16 backbone, which returns a 512-dim feature vector. QDHF learns a latent projection from 512-d to 2-d. To gather online human feedback, we use DreamSim with the DINO-ViT-B/16 backbone. The DreamSim model is trained on the NIGHTS dataset, which consists of 20k synthetic image triplets annotated with human judgments as labels. For QDHF, we use 10000 judgments of predicted human feedback. The objective is the CLIP score of the image and the text prompt. The text prompt is also input to the Stable Diffusion model to condition the generation towards more relevant content.

\section{Additional Results} \label{sec:aresults}

In this section, we include more experimental results to for further analysis of the proposed method.

\subsection{Robotic Arm} \label{sec:arobotic}

\paragraph{Alignment between learned and ground truth diversity.}

\begin{figure}[t]
  \vskip 0.1in
  \begin{center}
    \includegraphics[width=\linewidth]{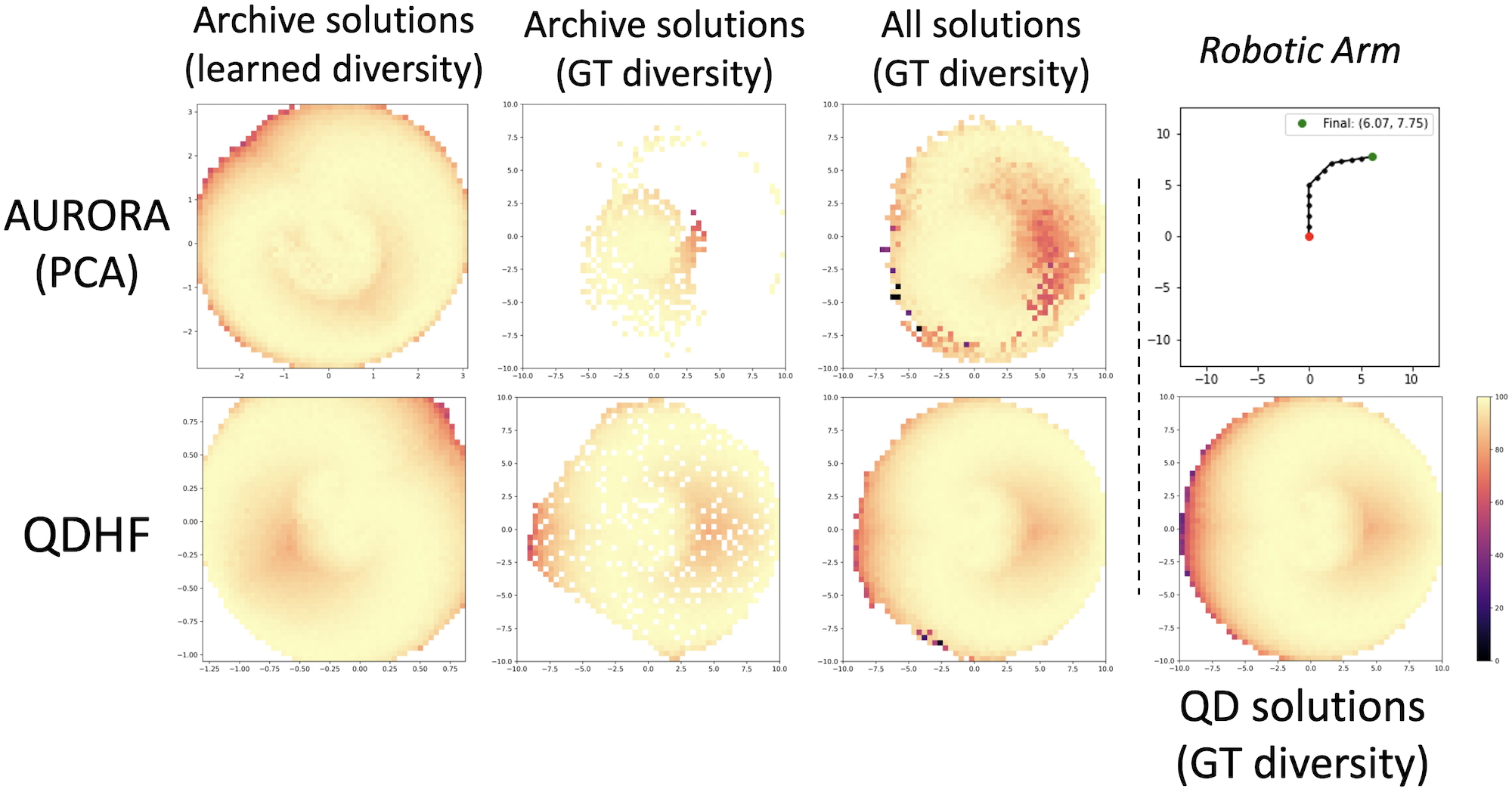}\\
    \caption{Visualization of the solution archives on different diversity spaces for the robotic arm task. ``GT'' stands for ground truth. Each point on the heatmap is a solution with its objective value visualized in color. QDHF fills up the archives with more solutions than AURORA, and more accurately learns the scale of the ground truth diversity metrics.}
    \label{fig:arm_heat}
  \end{center}
  \vskip -0.1in
\end{figure}

We evaluate the alignment of diversity metrics derived by QDHF with the underlying ground truth diversity metrics. In Fig.~\ref{fig:arm_heat}, the archives of QDHF and AURORA-Inc (PCA) are visualized. Both AURORA and QDHF appear to effectively learn a diversity space reflective of the ground truth. However, QDHF exhibits enhanced capability in discerning the relative distances between solutions. This efficacy stems from QDHF's ability to better align its learned diversity space with the ground truth diversity, especially concerning the scales on each axis.

\begin{figure*}[t]
  \vskip 0.1in
  \begin{center}
    \includegraphics[width=.84\linewidth]{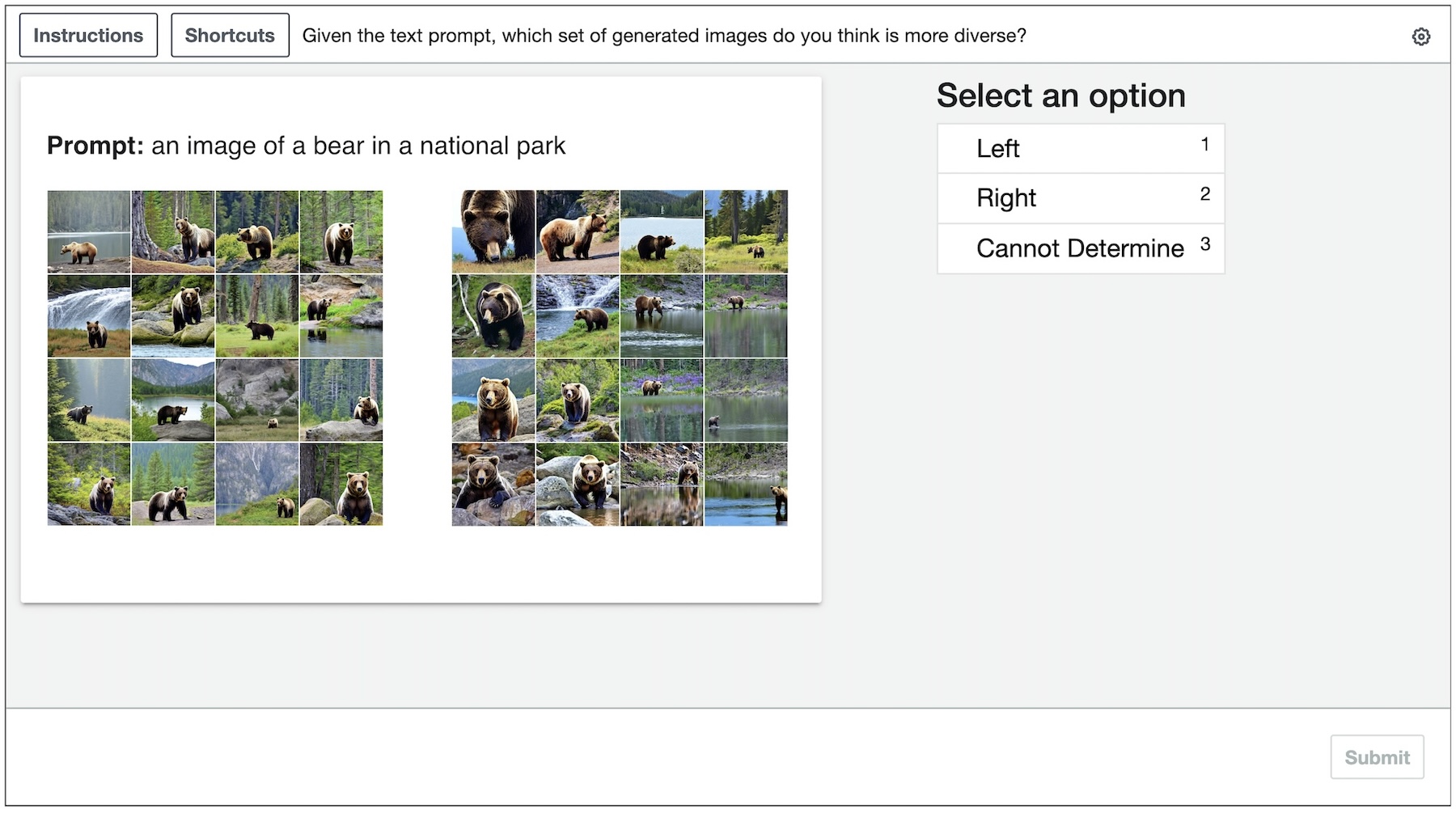}\\
    \caption{Example of the human evaluation interface on Amazon Mechanical Turk.}
    \label{fig:turk}
  \end{center}
  \vskip -0.1in
\end{figure*}

\begin{table*}[t]
  \caption{Results for robotic arm with noisy labels. We report the QD score (normalized to a scale of 0-100) and coverage for ``All Solutions'' (solutions found throughout training) and ``Archive Solutions'' (solutions in the final archive). We additionally report the performance of QDHF at different noise levels, where 5\% Noise means 5\% of the preference labels being randomly sampled are flipped. QDHF's performance is not significantly affected at the low noise level (5\%), and still outperforms QDHF-Base (with perfect data) and AURORA at the highest noise level of 20\%.}
  \label{tb:arm_noise}
  \vskip 0.1in
  \begin{center}
    \begin{small}
      \begin{tabular}{lcccc}
        \toprule
                          & \multicolumn{2}{c}{All Solutions} & \multicolumn{2}{c}{Archive Solutions}                                                     \\
        \cmidrule(lr){2-3}\cmidrule(lr){4-5}
        Methods           & QD Score                          & Coverage                              & QD Score                & Coverage                \\
        \midrule
        AURORA-Pre (AE)   & $38.5 \pm 12.7$                   & $56.2 \pm 6.8$                        & $14.3 \pm 6.3$          & $20.3 \pm 6.4$          \\
        AURORA-Inc (AE)   & $53.0 \pm 9.4$                    & $63.2 \pm 6.2$                        & $17.6 \pm 4.1$          & $18.9 \pm 4.1$          \\
        AURORA-Pre (PCA)  & $38.4 \pm 13.6$                   & $54.1 \pm 9.0$                        & $14.2 \pm 4.4$          & $19.3 \pm 5.5$          \\
        AURORA-Inc (PCA)  & $45.9 \pm 6.4$                    & $59.0 \pm 3.7$                        & $18.3 \pm 3.4$          & $19.0 \pm 3.5$          \\
        \midrule
        QDHF-Base         & $54.5 \pm 4.3$                    & $62.7 \pm 2.7$                        & $31.8 \pm 4.5$          & $34.1 \pm 4.2$          \\
        QDHF              & $\mathbf{72.5 \pm 0.9}$           & $\mathbf{77.3 \pm 1.2}$               & $\mathbf{56.4 \pm 0.9}$ & $\mathbf{59.9 \pm 0.9}$ \\
        \midrule
        QDHF (5\% Noise)  & $72.4 \pm 0.8$                    & $76.9 \pm 1.0$                        & $55.6 \pm 0.7$          & $58.7 \pm 0.5$          \\
        QDHF (10\% Noise) & $68.3 \pm 4.7$                    & $73.8 \pm 4.2$                        & $44.5 \pm 11.5$         & $47.2 \pm 12.0$         \\
        QDHF (20\% Noise) & $62.6 \pm 5.1$                    & $70.6 \pm 2.4$                        & $35.4 \pm 10.1$         & $37.9 \pm 9.9$          \\
        \midrule
        QD-GT             & $74.8 \pm 0.2$                    & $79.5 \pm 0.3$                        & $74.8 \pm 0.2$          & $79.5 \pm 0.3$          \\
        \bottomrule
      \end{tabular}
    \end{small}
  \end{center}
  \vskip -0.2in
\end{table*}

\paragraph{Noisy feedback.}

To investigate the impact of noise in human feedback data on the performance of QDHF, we conducted an ablation study on the robotic arm task. In this study, we manually injected noise into the feedback data, which was simulated from a ground truth diversity metric. We selected three noise levels—5\%, 10\%, and 20\%—and at each level, a corresponding percentage of the feedback data, randomly sampled from the entire dataset, was flipped. This simulates potential bias or errors by human annotators during the annotation process.

The results, as presented in Table~\ref{tb:arm_noise}, indicate that QDHF's performance is not significantly affected at the lower noise level of 5\%. Although QDHF's performance declines at higher noise levels, it still outperforms the baseline method (with perfect data) and all variants of AURORA, even at the highest noise level of 20\%. These findings validate QDHF's robustness against potential noise in the feedback data from humans.

\subsection{Latent Space Illumination} \label{sec:alsi}

We include more detailed experimental setups and results for the LSI task.

\paragraph{Quantitative results on compositional prompts.}

We include the quantitative evaluations of results on compositional prompts in Table~\ref{tab:lsi_quan_comp}, which are summarized over 100 text prompts. QDHF has a similar CLIP score to SD (Best-of-n), but much higher mean and standard deviation of pairwise distance, which indicates that QDHF is able to generate more diverse solutions while maintaining the high-quality. We can see that even though the CLIP scores is high, the issue of attribute leakage or interchanged attributes still exists in the results. However, QDHF is able to mitigate this issue by exploring more diverse responses, which results in a higher chance of approaching the correct response. For a more accurate evaluation of correctness, we rely on human evaluation as shown in Table~\ref{tab:lsi_qual}.

\begin{table}[t]
  \caption{Quantitative Results for LSI (compositional prompts). We report the CLIP score and DreamSim pairwise distance (mean and std.) as quantitative metrics. QDHF demonstrates competitive quality to Stable Diffusion (Best-of-n) measured by CLIP score, and significantly better diversity measured by pariwise distance.}
  \label{tab:lsi_quan_comp}
  \vskip 0.1in
  \begin{center}
    \begin{small}
      \begin{tabular}{lccc}
        \toprule
        Method         & CLIP Score       & Mean PD          & Std. PD          \\
        \midrule
        SD (Best-of-n) & $71.58$          & $0.410$          & $0.088$          \\
        QDHF           & $\mathbf{71.77}$ & $\mathbf{0.543}$ & $\mathbf{0.128}$ \\
        \bottomrule
      \end{tabular}
    \end{small}
  \end{center}
  \vskip -0.1in
\end{table}

\paragraph{User study details.} The LSI experiments are conducted with singular prompts and compositional prompts. We use human evaluation to assess the results through a ``blind'' user study, where the users are unaware of the origin of each output when making the decision.
For singular prompts, we use 100 prompts and each prompt produces results of two 4x4 grids of images for SD (Best-of-N) and QDHF, respectively.
We evaluate the results by asking the users to answer two questions: (1) which set of generated images is more preferred, and (2) which set of generated images is more diverse.
The study is conducted on Amazon Mechanical Turk, where each evaluation task is created as a Human Intelligence Task (HIT).
For each prompt, we seek 50 evaluations from different users, and a user is qualified if they have a HIT Approval Rate over $95\%$. We follow the Amazon MTurk guidelines to fairly compensate the annotators for \$0.01 per evaluation per text prompt. An example of the MTurk annotation interface is shown in Fig.~\ref{fig:turk}.

For compositional prompts, we follow a similar human evaluation procedure, and the question is: which set of generated images has the most accurate response.
We also conduct experiments on 100 text prompts, sampled from the MCC-250 benchmark~\citep{bellagente2023multifusion}, with 50 evaluations per prompt.
The comparisons are done on 3x3 grids of images since the evaluation target is on correctness rather than diversity.

\paragraph{More qualitative results on singular prompts.}

\begin{figure*}[t]
  \vskip 0.1in
  \begin{center}
    \includegraphics[width=\linewidth]{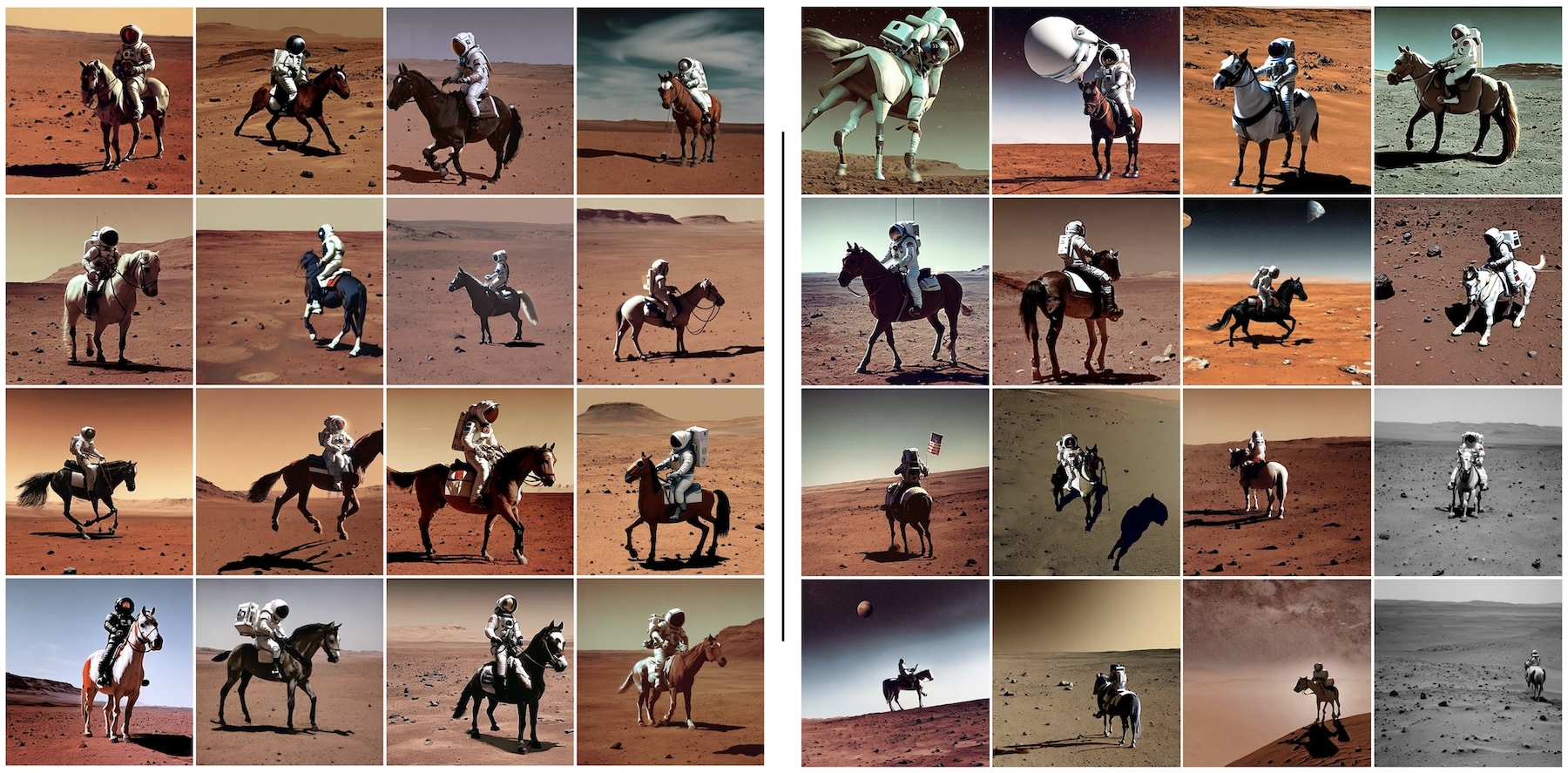}\\
    \caption{Qualitative results for singular prompt: ``a photo of an astronaut riding a horse on mars''.}
    \label{fig:lsi1}
  \end{center}
  \vskip -0.1in
\end{figure*}

\begin{figure*}[t]
  \vskip 0.1in
  \begin{center}
    \includegraphics[width=\linewidth]{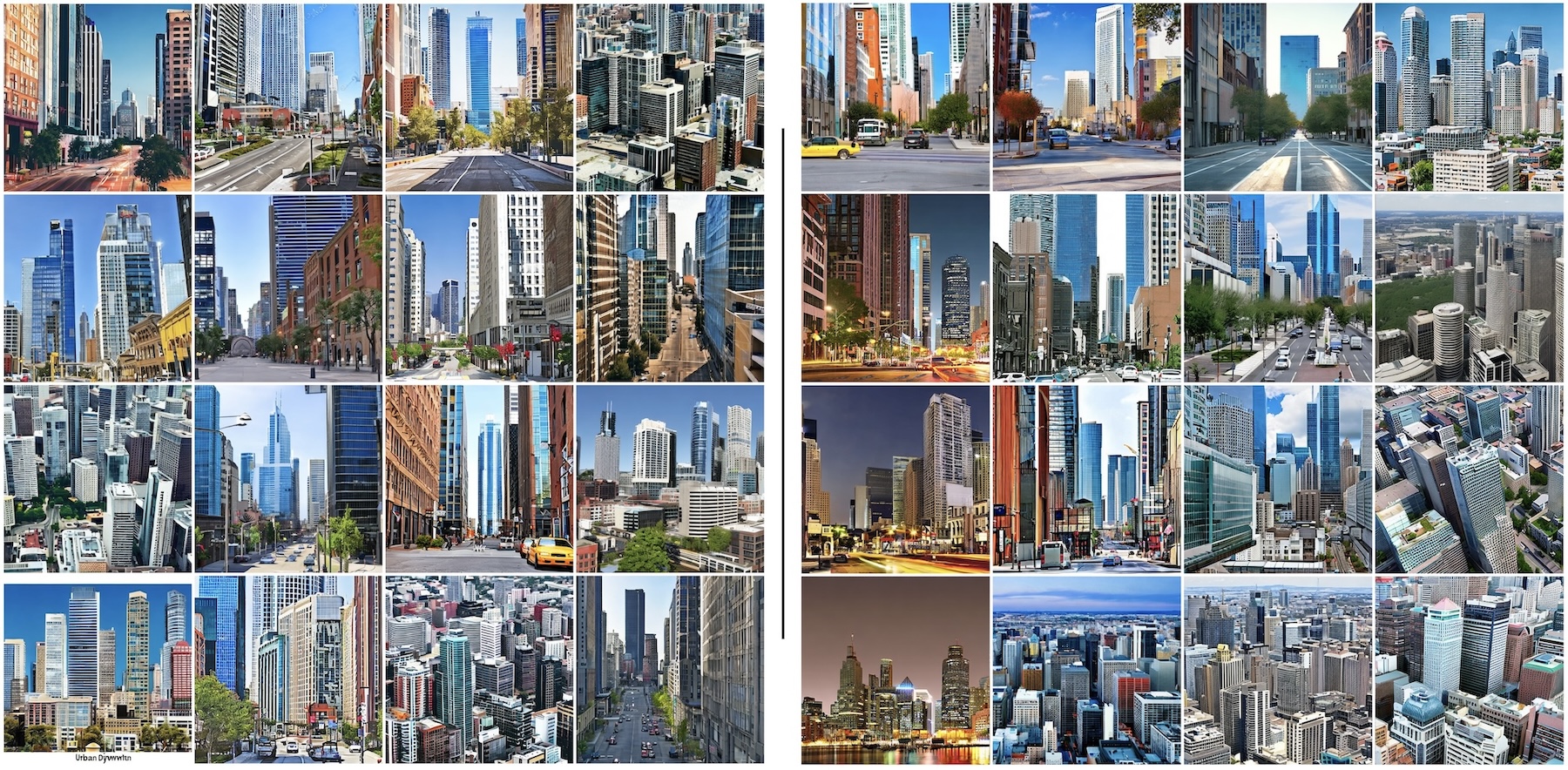}\\
    \caption{Qualitative results for singular prompt: ``an image of urban downtown''.}
    \label{fig:lsi6}
  \end{center}
  \vskip -0.1in
\end{figure*}

We show more qualitative results of generated images with singular prompts in Fig.~\ref{fig:lsi1} to Fig~\ref{fig:lsi6}. For all the figures, the left 4x4 grid displays images with the highest CLIP scores from randomly generated images, and the right grid displays a uniformly sampled subset of QDHF solutions. Qualitatively, images generated by QDHF have more variations and show visible trends of diversity.

Notably, while QDHF significantly outperforms the baseline on most cases, we find that Fig~\ref{fig:lsi6} is a sub-performing case for QDHF. Although QDHF is able to generate both day and night scenes, the diversity between the scenes is not apparent. The most likely reason is that the preference model (DreamSim) does not generalize well to cases such as different appearances of cityscapes. We aim to solve the above issues in future work where human feedback needs to be collected in a more diverse and strategic way to facilitate better generalization of the preference model and thus improve the performance of QDHF.

Another interesting finding is that for prompt Fig~\ref{fig:lsi1}, while most users find QDHF results are more diverse, more than half of the users actually prefer the less diverse baseline results. According to the feedback from users, people may prefer less diverse but more content-focused results in some specific cases. The relationship between diversity and user preference under different use cases in generative AI applications remains an open question, and we look forward to exploring this topic in future work.

\begin{figure*}[t]
  \vskip 0.1in
  \begin{center}
    \includegraphics[width=\linewidth]{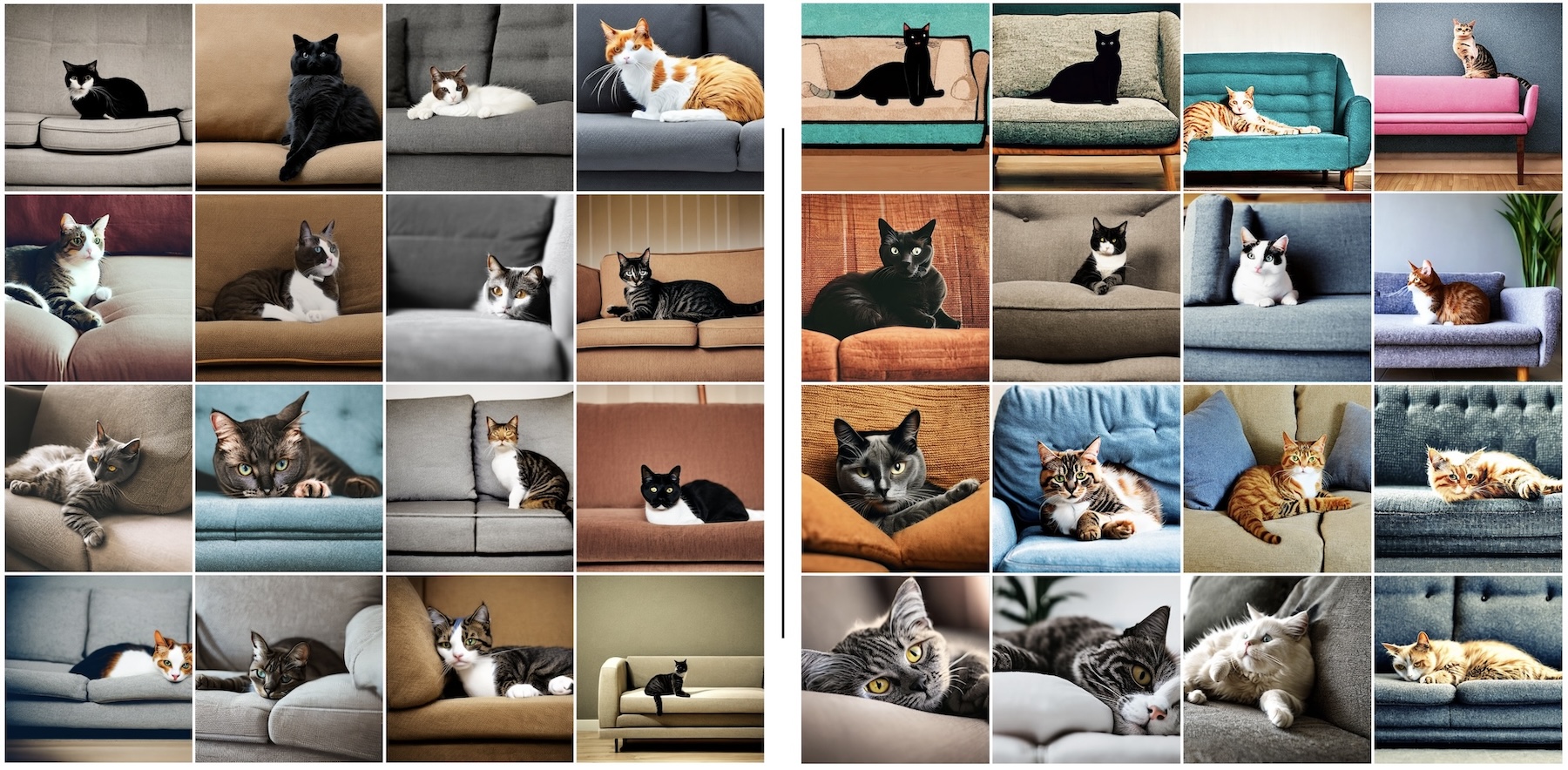}\\
    \caption{Qualitative results for singular prompt: ``an image of a cat on the sofa''.}
    \label{fig:lsi2}
  \end{center}
  \vskip -0.1in
\end{figure*}

\begin{figure*}[t]
  \vskip 0.1in
  \begin{center}
    \includegraphics[width=\linewidth]{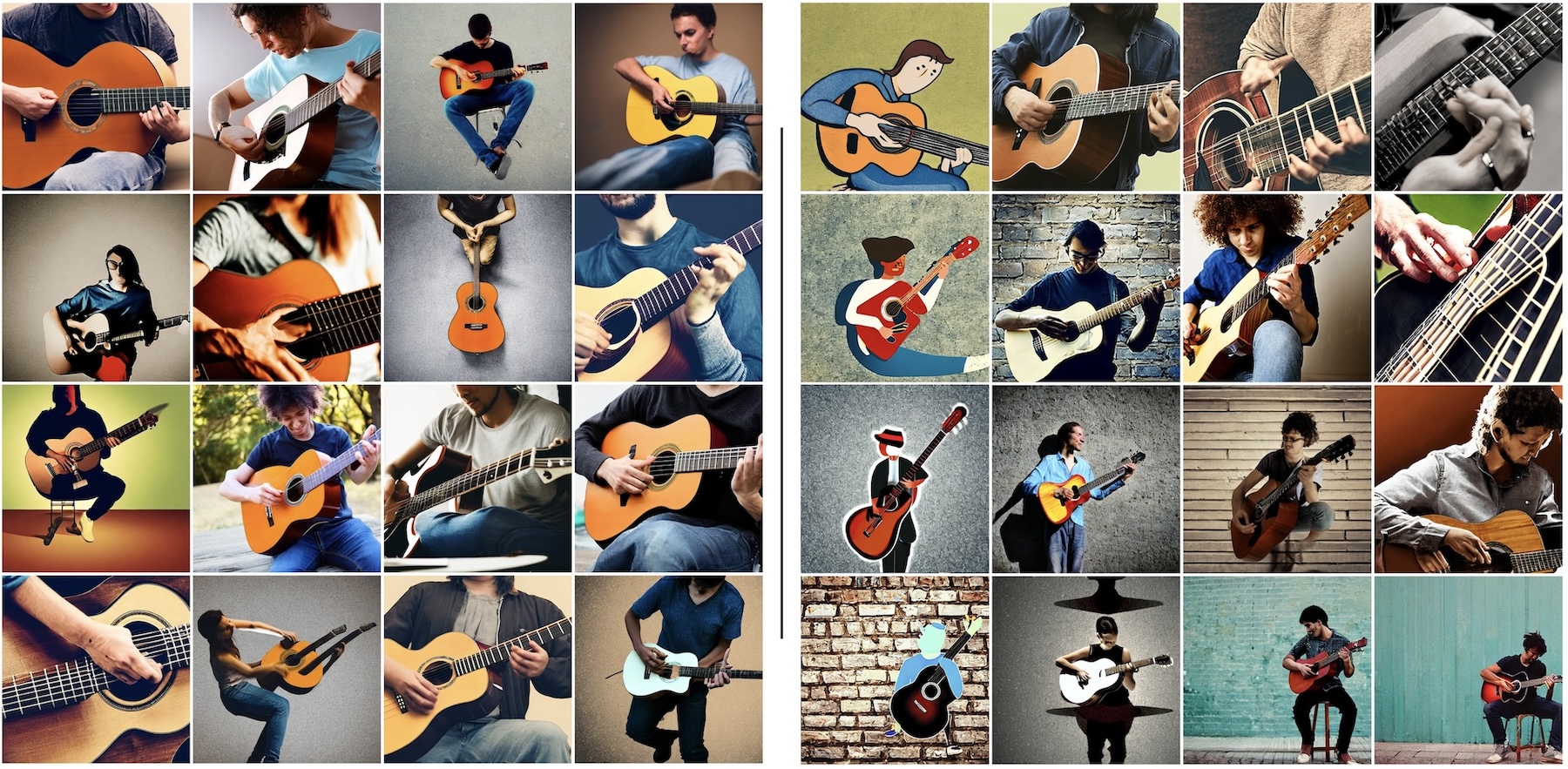}\\
    \caption{Qualitative results for singular prompt: ``an image of a person playing guitar''.}
    \label{fig:lsi4}
  \end{center}
  \vskip -0.1in
\end{figure*}

\paragraph{More qualitative results on compositional prompts.}

\begin{figure*}[t]
  \vskip 0.1in
  \begin{center}
    \includegraphics[width=.9\linewidth]{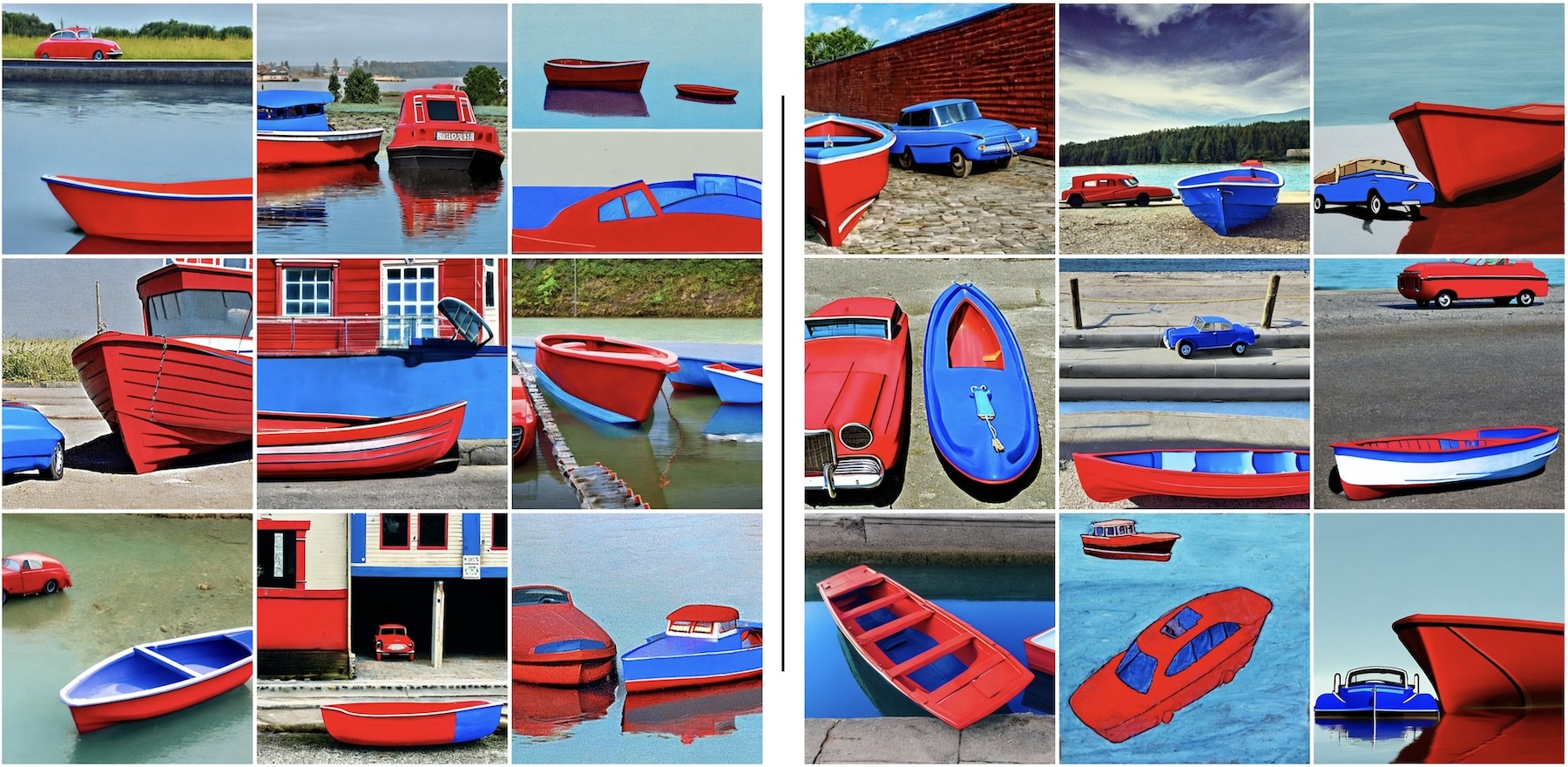}\\
    \caption{Qualitative results for compositional prompt: ``a red boat and a blue car''.}
    \label{fig:lsi_obj1}
  \end{center}
  \vskip -0.1in
\end{figure*}

\begin{figure*}[t]
  \vskip 0.1in
  \begin{center}
    \includegraphics[width=.9\linewidth]{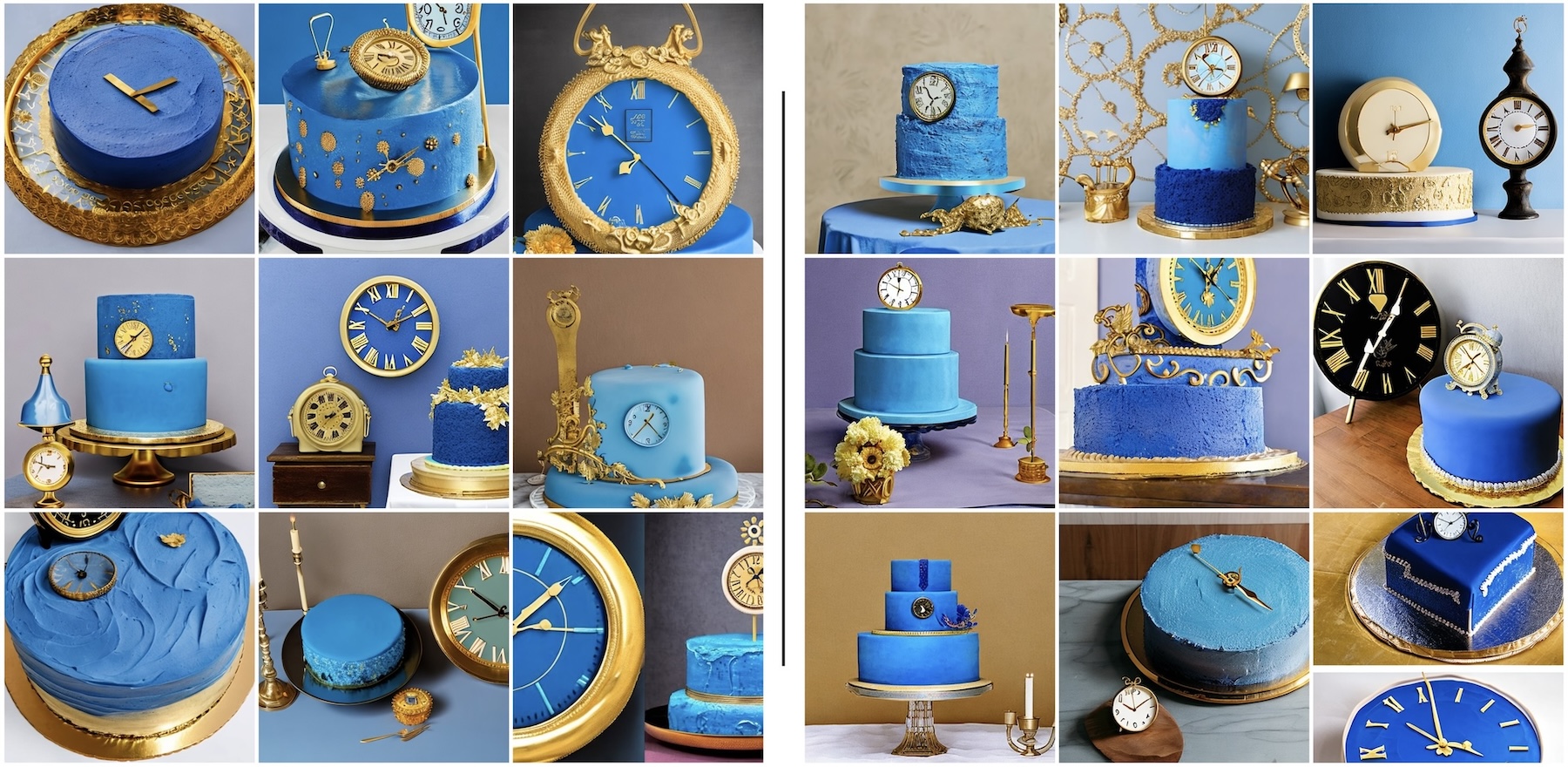}\\
    \caption{Qualitative results for compositional prompt: ``a blue cake and a gold clock''.}
    \label{fig:lsi_obj2}
  \end{center}
  \vskip -0.1in
\end{figure*}

We also include more qualitative results of generated images with compositional prompts in Fig.~\ref{fig:lsi_obj1} and Fig~\ref{fig:lsi_obj2}. For all the figures, the left 3x3 grid displays images with the highest CLIP scores from randomly generated images, and the right grid displays a uniformly sampled subset of QDHF solutions. Qualitatively, images generated by QDHF have more variations and show a higher frequency of correctness.

